\documentclass[lettersize,journal]{IEEEtran}
\usepackage{amsmath,amsfonts}
\usepackage{algorithmic}
\usepackage{algorithm}
\usepackage{array}
\usepackage{textcomp}
\usepackage{stfloats}
\usepackage{url}
\usepackage{verbatim}
\usepackage{graphicx}
\usepackage{cite}
\usepackage{multirow}
\usepackage{pifont}
\usepackage{subcaption}
\usepackage{color}
\usepackage{cuted}
\usepackage{capt-of}
\usepackage{booktabs}
\usepackage{amssymb}
\usepackage{adjustbox}
\usepackage{makecell} 
\usepackage{pdfpages}

\begin{document}
\title{Extreme Two-View Geometry From Object Poses with Diffusion Models}

\author{Yujing Sun*,
		Caiyi Sun*,
		Yuan Liu*,
		Yuexin Ma,
		Siu Ming Yiu
        \thanks{* denotes equal contribution}
\thanks{Yujing Sun, Yuan Liu, and Siu Ming Yiu are with the University of Hong Kong (Emails: \{yjsun@cs, yuanly@connect, smyiu@cs\}.hku.hk). Caiyi Sun is with Beijing Institute of Technology (Email: scy639@outlook.com). Yuexin Ma is with Shanghai Technology University (Email: mayuexin@shanghaitech.edu.cn).  }
\thanks{Manuscript received April 19, 2021; revised August 16, 2021.}}

\markboth{Journal of \LaTeX\ Class Files,~Vol.~14, No.~8, August~2021}%
{Shell \MakeLowercase{\textit{et al.}}: A Sample Article Using IEEEtran.cls for IEEE Journals}


\maketitle

\begin{strip}
\vspace{-7em}
 \begin{center}  
    \includegraphics[width=0.47\textwidth]{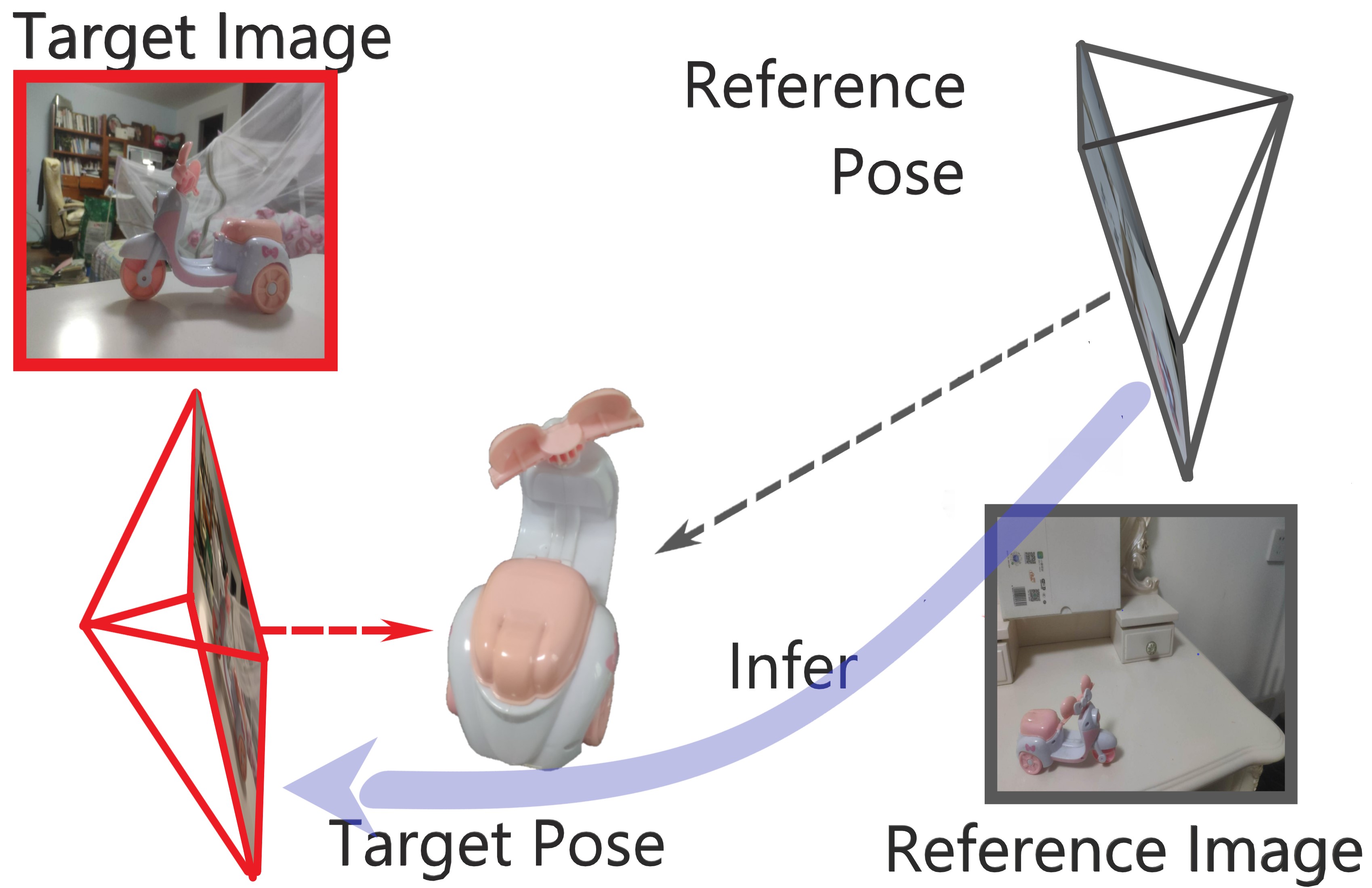}
    \hspace{+2em}
    \includegraphics[width=0.47\textwidth]{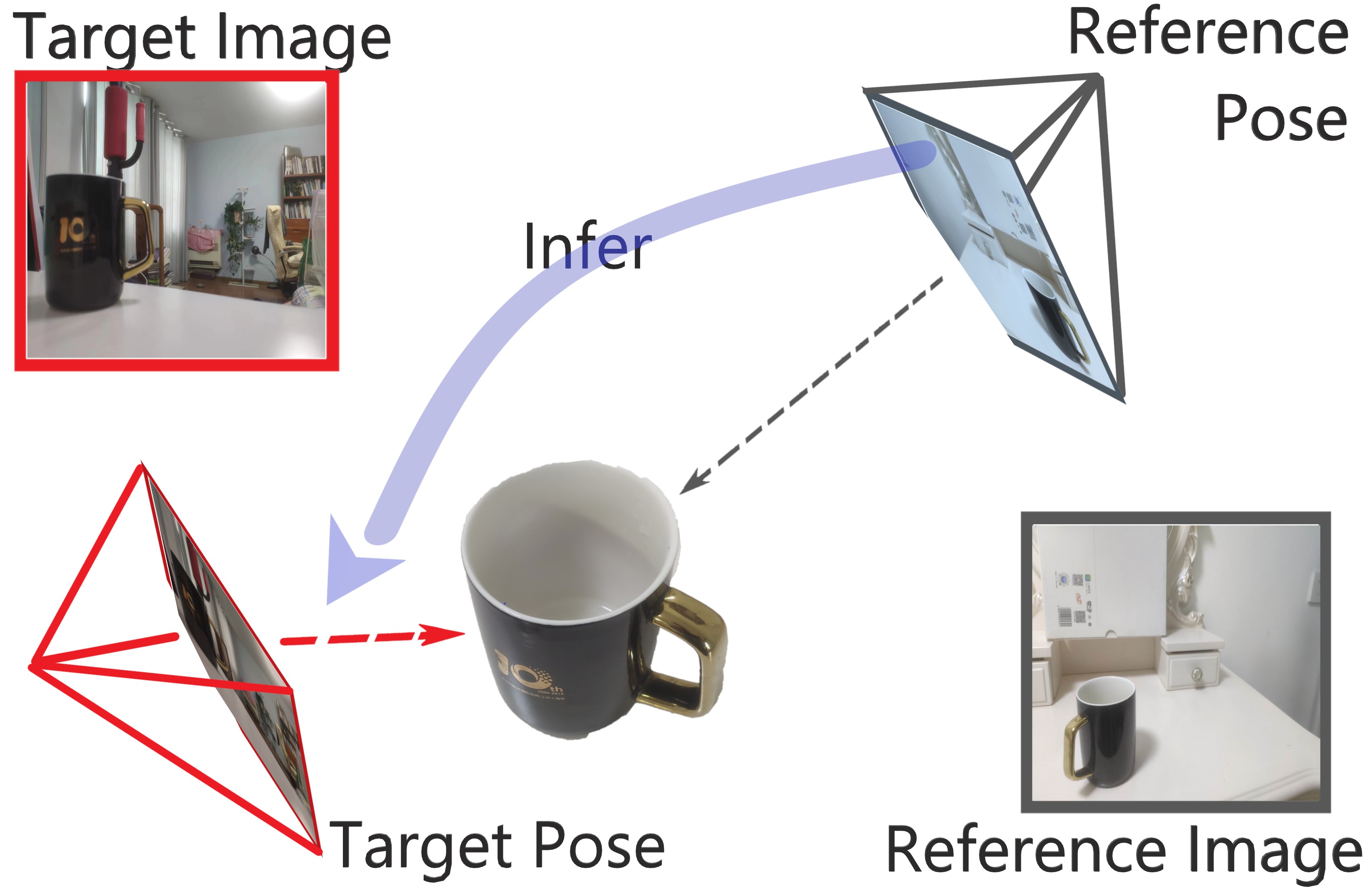}
    \end{center}
    \vspace{+2em}
    \begin{center}
    \includegraphics[width=0.47\textwidth]{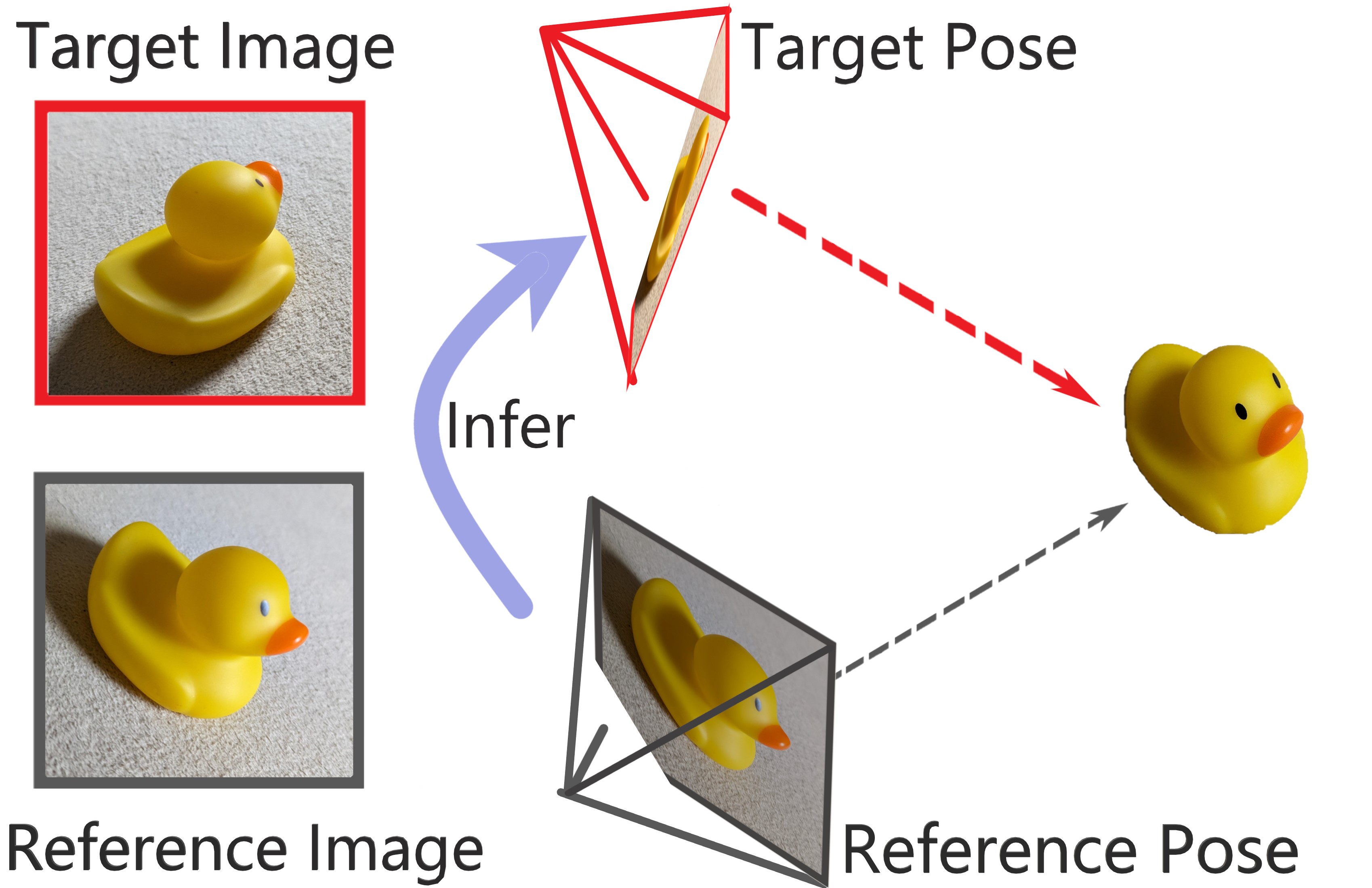}
    \hspace{+2em}
     \includegraphics[width=0.47\textwidth]{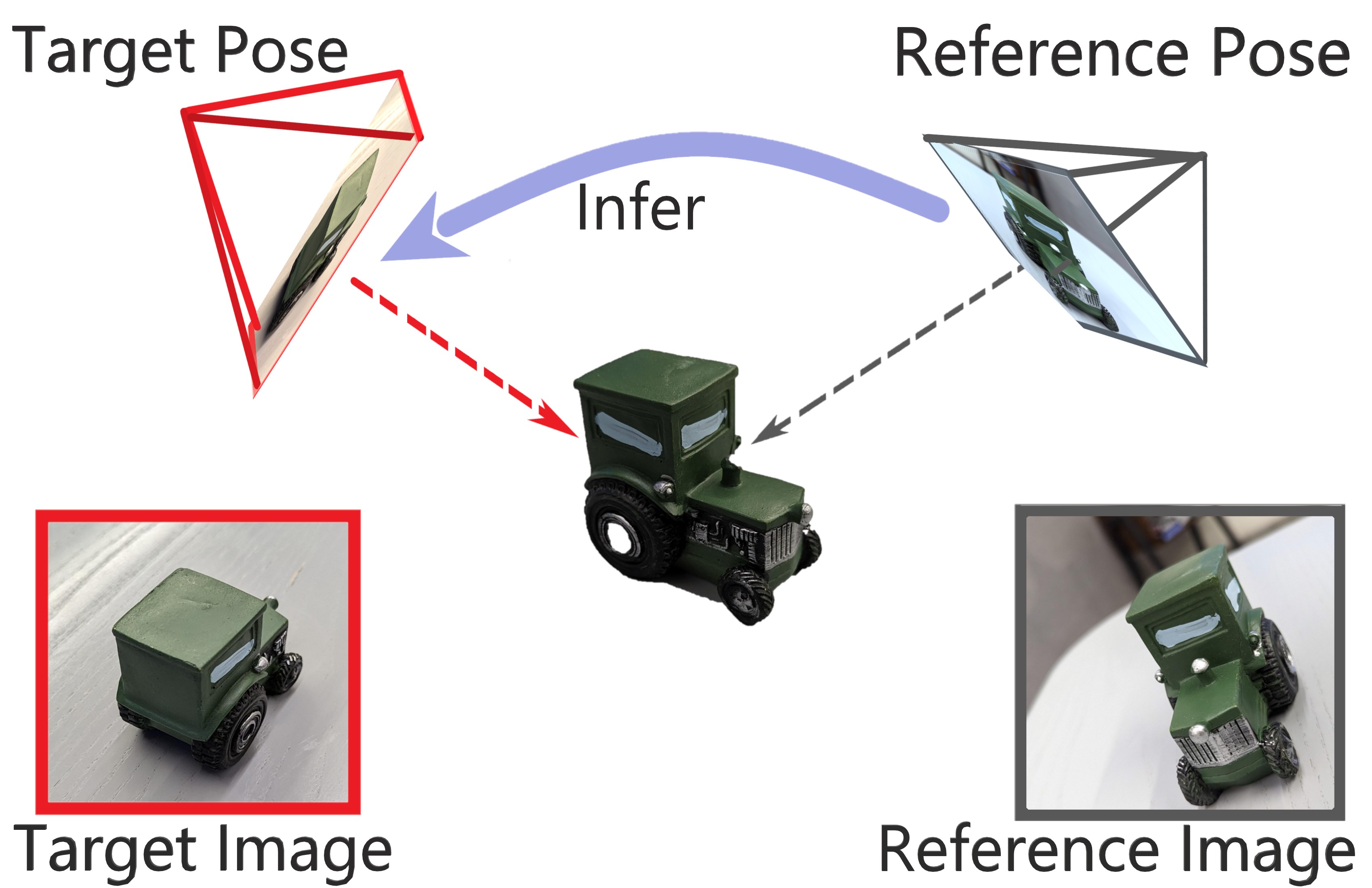}   
	\end{center}   
 \captionof{figure}{Our method can accurately infer the extreme relative camera pose of two images containing a co-visible object even without any overlap regions for correspondence estimation. Our method is based on a diffusion generative model to hallucinate the unseen sides of the object and match the hallucinated images with query images to estimate relative camera poses. The estimated extreme camera poses can be used in downstream applications, e.g. visual odometry.
 }
\label{fig:teaser}
\end{strip}

\begin{abstract}

Human has an incredible ability to effortlessly perceive the viewpoint difference between two images containing the same object, even when the viewpoint change is astonishingly vast with no co-visible regions in the images. This remarkable skill, however, has proven to be a challenge for existing camera pose estimation methods, which often fail when faced with large viewpoint differences due to the lack of overlapping local features for matching. In this paper, we aim to effectively harness the power of object priors to accurately determine two-view geometry in the face of extreme viewpoint changes. In our method, we first mathematically transform the relative camera pose estimation problem to an object pose estimation problem. Then, to estimate the object pose, we utilize the object priors learned from a diffusion model Zero123~\cite{zero123} to synthesize novel-view images of the object. The novel-view images are matched to determine the object pose and thus the two-view camera pose.
In experiments, our method has demonstrated extraordinary robustness and resilience to large viewpoint changes, consistently estimating two-view poses with exceptional generalization ability across both synthetic and real-world datasets. Code will be available at https://github.com/scy639/Extreme-Two-View-Geometry-From-Object-Poses-with-Diffusion-Models.

\end{abstract}


\section{Introduction}
\label{sec:intro}


Relative camera pose estimation is a fundamental task in the realm of computer vision, with numerous applications spanning various cutting-edge techniques. However, estimating the relative poses of two views with extreme viewpoint changes presents a formidable challenge, especially when the co-visible regions of the two views are textureless. This difficulty stems from the lack of distinctive features to establish reliable correspondences between the two views, which is essential for accurate pose estimation. As the demand for high-quality 3D reconstruction~\cite{mildenhall2021nerf}, augmented reality~\cite{liu2022gen6d}, and other computer vision applications continue to grow, addressing this challenge is paramount for unlocking the full potential of these advanced techniques and pushing the boundaries of what is possible in the field of computer vision.

Though the widely adopted feature matching has difficulty in estimating camera poses of two views with extreme viewpoint changes, humans possess a remarkable ability to estimate extreme two-view poses when a common object is present in both views. This innate skill suggests that leveraging such object priors could hold the key to designing a new algorithm for two-view pose estimation, overcoming the limitations of current feature-matching techniques, and enabling more accurate and robust estimation of camera poses in challenging scenarios. 

How to utilize such object priors remains an open problem. Recent works RelPose~\cite{relpose2022}, RelPose++~\cite{relpose++2023} and SparsePose~\cite{sparepose2023} utilize the object prior by training a transformer on the CO3D dataset~\cite{co3D2021} to regress or score the relative pose of two views of the same object. Though some promising results are achieved, we find that using the transformer to model the object prior leads to less generalization ability, especially on datasets of different domains from the CO3D dataset. 



Recently, diffusion models~\cite{rombach2022high} that have been trained on billion-scale datasets~\cite{schuhmann2022laion} have shown to be capable of generating high-quality images. Such a generation ability indicates these diffusion models learn meaningful priors over natural images. Based on the pretrained diffusion models, many recent methods~\cite{zero123,liu2023syncdreamer,long2023wonder3d,shi2023zero123++} have finetuned to generate diverse and high-quality images of arbitrary objects. This means these models encode robust object priors, which are much more generalizable than the transformer trained only on the CO3D dataset. Thus, these diffusion models are very promising tools for solving the extreme two-view camera pose estimation with object prior.

\begin{figure}[b]
    \centering
    \includegraphics[width=0.48\textwidth]{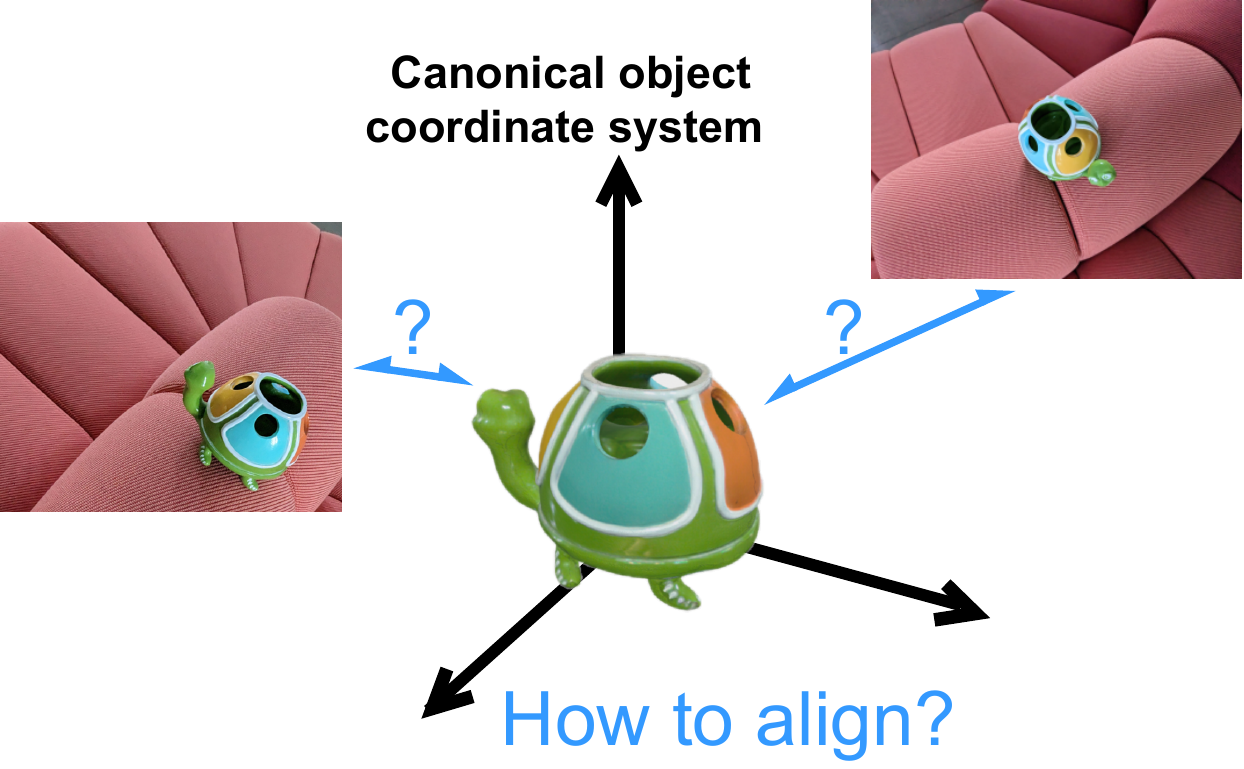}
    \caption{Challenges in applying the object prior from diffusion models, e.g. Zero123~\cite{zero123}, to relative pose estimation. On one hand, input images may not look at the object while Zero123 and common object pose estimations require the object to be located at the image center. On the other hand, Zero123 implicitly defines a canonical object coordinate inside, which brings difficulty in aligning the input images to this implicit canonical coordinate object system.}
    \label{fig:how_to_align}
\end{figure}
Since these diffusion models are designed for generating images of objects, how to utilize them in the two-view pose estimation still remains unexplored. These diffusion models all assume that the image is strictly looking at the object when generating novel-view images for the object, as shown in Fig.~\ref{fig:how_to_align}. However, the input images may just contain the object but not strictly look at the object. Even worse, these diffusion models implicitly define a canonical object coordinate system in the model, which makes it even more difficult to find the poses and intrinsic matrices of the generated images in the canonical object coordinate system. Meanwhile, we have to derive the relative transformation between the camera coordinate system and the canonical object coordinate system used in diffusion models. 


In this paper, we aim to build a novel framework, as shown in Fig.~\ref{fig:teaser}, to estimate extreme two-view camera poses by utilizing the object prior from a diffusion model. In order to utilize the diffusion mode in a pose estimation framework, we first propose a new formulation of two-view pose estimation by transforming the camera pose estimation problem into an object pose estimation problem. 
Then, on this object pose estimation problem, we generate multiple images of this object captured from different viewpoints by the Zero123~\cite{zero123} model. On these generated images, we propose a novel and effective way to define their object poses and intrinsic matrices. Then, we match the other input image against the generated images to determine the plausible object poses, similar to Gen6D~\cite{liu2022gen6d}. Finally, the determined object poses are then mapped back to determine the two-view camera pose of the input image pair. 

Extensive experiments demonstrate that our method is able to predict accurate relative camera poses on image pairs with extreme viewpoint changes. On both synthetic and real datasets, our method outperforms baseline matching-based methods~\cite{sarlin2020superglue,sun2021loftr} and the transformer-based pose regressor~\cite{relpose2022, relpose++2023} by a large margin. Our method shows strong generalization ability on the in-the-wild datasets and can be used in improving the closure optimization of visual odometry (VO).

In summary, the contribution of our paper is as follows:

\begin{itemize}
\item We propose a novel pose estimation algorithm for image pairs with extreme viewpoints change, based on diffusion models and image matching. 
\item To enable the utilization of object priors in diffusion models, we mathematically transform the relative pose estimation problem into an object pose estimation problem. 
\item The proposed framework significantly improves the accuracy of extreme two-view pose estimation on both synthetic and real datasets and shows promising results in combination with a VO method.  
\end{itemize} 

\section{Literature Review}
\label{sec:lit}
\textbf{Feature Matching Based Pose Estimation} \quad
In the scenarios of estimating poses from sets of images or video streams~\cite{longuet1981computer,nister2004efficient}, traditional methods typically revolve around finding correspondences between specifically designed local features~\cite{lowe2004distinctive,bay2006surf,tola2009daisy}. 
Recently, with the development of deep learning techniques, neural networks have been employed and explored to boost the matching accuracy and robustness~\cite{liu2010sift,choy2016universal,sarlin2020superglue,sun2021loftr}. 
Such matching-based pose estimation has been widely used in the downstream applications, including Structure-from-Motion (SfM)~\cite{schonberger2016structure}, Multi-View Stereo (MVS)~\cite{schonberger2016pixelwise} and Simultaneously localization and mapping (SLAM)~\cite{mur2015orb,mur2017orb,campos2021orb}.
However, in challenging scenarios where images are sparsely captured, these methods often struggle to accurately estimate poses due to the limited availability of effective features. 
Consequently, these approaches are not well-suited or perform poorly under such sparse-views setting.

\textbf{Dense/multi-Views Based Pose Estimation} \quad
Another main class of pose estimation is based on multiple or dense reference views. To achieve the goal, different priors are employed, such as category-specific knowledge,~\cite{kanazawa2019learning,kocabas2020vibe,ma2022virtual,usman2022metapose,wen2022disp6d}, and temporal locality for SLAM and VO applications~\cite{wang2017deepvo,yang2020d3vo,teed2021}.

More recently, category-agnostic priors~\cite{li2017deepim,zhao2022fusing,liu2022gen6d,sun2022onepose,he2022oneposeplusplus,10350449_PoseMatcher,zhao2023locposenet} have been presented to generalize beyond object categories and achieved improved performance on the dense-views setting.
BundleTrack~\cite{wen2021bundletrack} and BundleSDF~\cite{bundlesdfwen2023} can perform object pose tracking given a monocular RGBD video sequence.
And PF-LRM~\cite{wang2023pf} performed 3D reconstruction with joint predictions of pose and shape. 
Some methods~\cite{10350892_NeRF_Pose,yenchen2021inerf,park2019latentfusion} also performed 3D reconstruction at first, which needs multiple input images.
However, methods falling in this category rely on the dense input reference views to provide enough objective priors, which cannot be obtained with a sparse-views setting. 

\textbf{Sparse-Views Based Pose Estimation} \quad
The proposed method is most relevant to literature belonging to this category. 
It is noticeable that the limited overlapping between views makes these setups much more challenging. 
Many methods have been proposed to estimate object pose based on sparse views but require additional object 3D model~\cite{Cai_2022_CVPR,9021977_CorNet,9423117_3DPoseLite,grabner2020geometric,okorn2021zephyr,9880271_OSOP,Xiao2019PoseFromShape,nguyen2022template} or depth maps~\cite{irshad2022centersnap,irshad2022shapo,he2022fs6d}.
When it comes to using only two RGB images for wide-baseline pose estimation, direct regression approaches~\cite{melekhov2017relative,rockwell20228} typically do not work well. 
More recently, by unifying optimizing pairs of relative rotations, RelPose~\cite{relpose2022} achieved satisfactory accuracy on the CO3D~\cite{co3D2021} dataset. 
Thereafter, transformer-based approach Relpose++~\cite{relpose++2023} and SparsePose~\cite{sparepose2023}  further achieved improved accuracy.
NOPE~\cite{nguyen2023nope} achieved the same goal via discriminative embedding prediction with UNet.
GigaPose~\cite{nguyen2023gigapose} also presented to estimate objects posed with a single image, but targeted CAD-like objects, while our method makes no assumptions on object types.
Meanwhile, E. Arnold et al.~\cite{arnold2022map} proposed a map-free re-localization with one single photo of a scene to enable instant, metric-scaled re-localization, which is also relevant literature.
However, most of the above works conduct experiments (training and testing) within the same dataset. In experiments, we observe that their performance dropped extremely when applying the pre-trained model to other datasets, indicating a limited generalization ability.
Compared with previous sparse-views or single-view-based approaches, the proposed framework demonstrates much stronger generalization capability and greatly outperforms them on in-the-wild datasets. 

It is notable that another mainstream on singe-view pose estimation assumes a fixed set of categories, such as humans~\cite{mehta2017vnect,kanazawa2018end} and also other predefined object categories~\cite{chang2015shapenet}. 
But our method is category-agnostic and depends on no category-related priors.
Meanwhile, there are also works related to the sparse-views-based pose estimation but with a rather different objective~\cite{jiang2022few}.  

\textbf{Diffusion Models} \quad
Image Diffusion Models~\cite{ho2020denoising,rombach2022high,mou2023t2i,zhang2023adding,zero123} have been extensively employed in image generation tasks, leveraging neural networks to denoise images through the estimation and removal of noise values that are blended into the image. 
These models gradually remove noise from pure noise, resulting in the generation of clear and high-quality images. 
In order to enhance efficiency and stability, latent diffusion models~\cite{rombach2022high,zero123} focused on denoising the latent representations of the image, and thereby achieved reducing model complexity. 
Recent advancements~\cite{mou2023t2i,zhang2023adding}, have enabled these models to be conditioned on additional inputs and provided improved control over the generated images.
One notable example of a latent diffusion model is Stable Diffusion~\cite{rombach2022high}, which achieves high-quality image generation by training on an extensive dataset of image-text pairs. 
Most recently, by fine-tuning Stable Diffusion~\cite{rombach2022high} on a collection of rendered pose-annotated images~\cite{deitke2023objaverse}, Zero-1-to-3~\cite{zero123} and Syndreamer~\cite{liu2023syncdreamer} achieved impressive results in generating novel views of an object by utilizing an input image of the object and information about its relative pose. 
In this paper, we attempt to effectively take advantage of diffusion models to generate useful object priors and to further guide the follow-up pose inference process.  
\section{Method}
Our method aims to estimate the relative camera pose of two images that have an extreme viewpoint change and a co-visible object as shown in Fig.~\ref{fig:teaser}. Given two images $I^{(1)}$ and $I^{(2)}$ with known intrinsic matrices $\mathbf{K}^{(1)}$ and $\mathbf{K}^{(2)}$ respectively, our target is to estimate the relative rigid transformation $[\mathbf{R}^{(12)};\mathbf{t}^{(12)}]$ with $\mathbf{x}_{c}^{(2)}=\mathbf{R}^{(12)}\mathbf{x}_{c}^{(1)} + \mathbf{t}^{(12)}$ where $\mathbf{x}_{c}^{(1)}$ and $\mathbf{x}_{c}^{(2)}$ are the points in the coordinate systems of two cameras respectively. Since the extreme viewpoint change here prevents us from building reliable correspondences to solve the camera pose, we estimate the relative camera poses utilizing object priors. 

\begin{figure*}[thbp] \centering
    \includegraphics[width=\textwidth ]{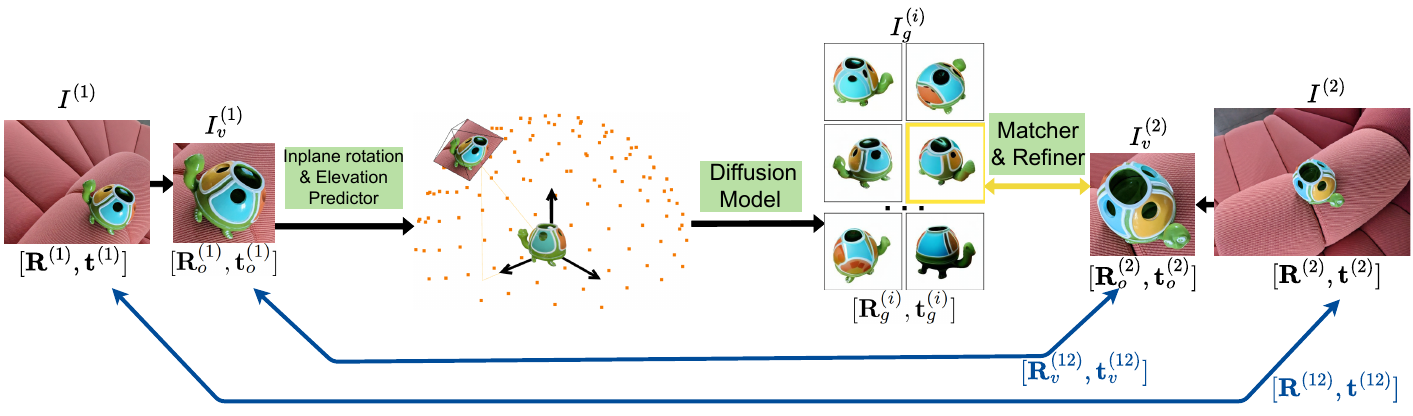}
    \caption{The overview of our pipeline.} \label{fig:overview}
\end{figure*}

The overview of our pipeline is shown in Fig.~\ref{fig:overview}. We first transform the input images into object-centric images $I_{v}^{(1)}$ and $I_{v}^{(2)}$ in Sec.~\ref{sec:obj_cen}. Then, we feed the image $I_{v}^{(1)}$ to a diffusion model~\cite{zero123} to generate a set of images on novel viewpoints $\mathcal{G}=\{I_{g}^{(i)} | i=1,2,3,...,N \}$ in Sec.~\ref{sec:diffusion}. Finally, we match the $I_{v}^{(2)}$ with the generated image set $\mathcal{G}$ to determine the relative pose between two images in Sec.~\ref{sec:matching}.
In the following, we first derive the underlying geometric relationship between the input image and the object-centric image.

\subsection{Object-Centric Images}
\label{sec:obj_cen}
\begin{figure}
    \centering
    \includegraphics[width=0.48\textwidth]{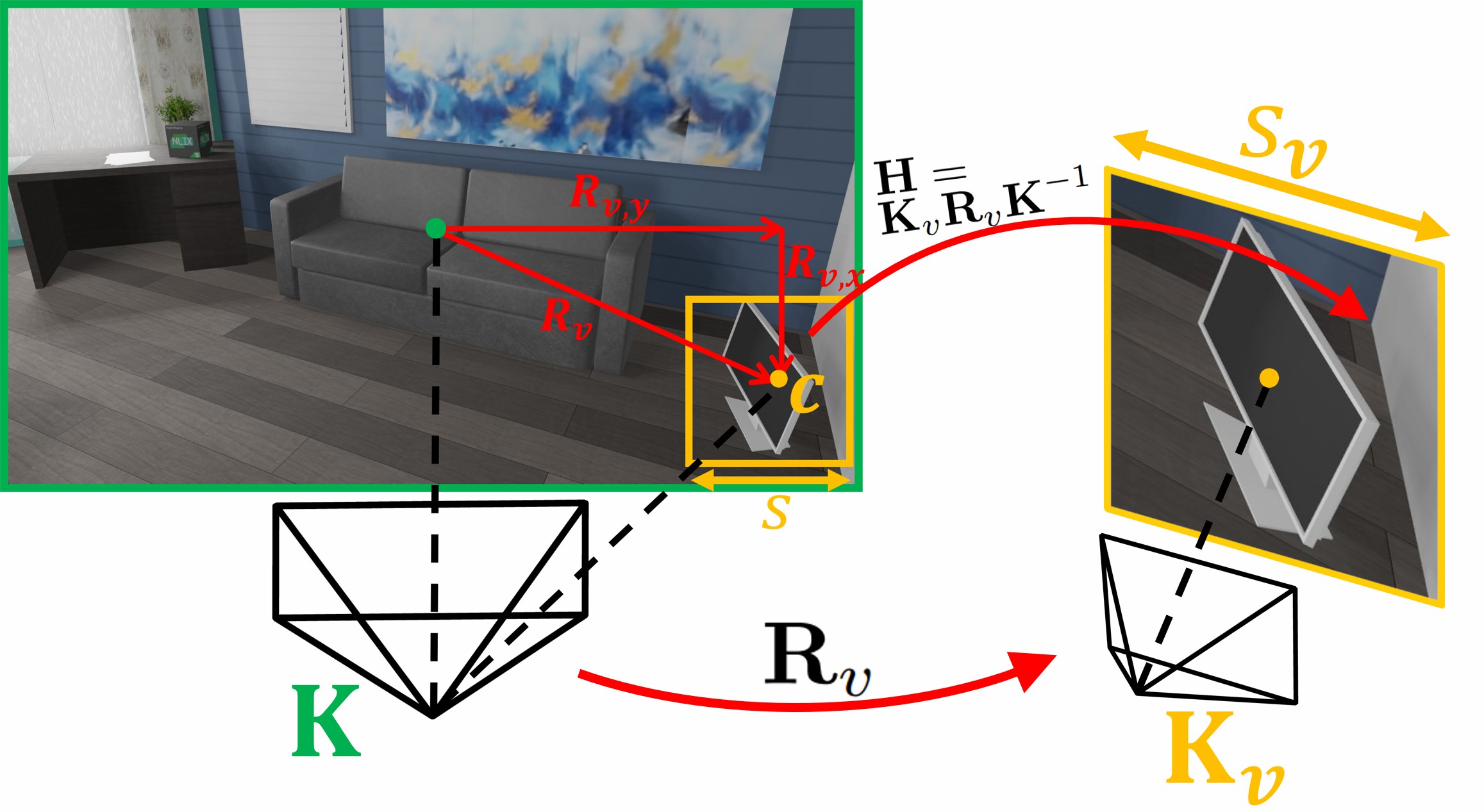}
    \caption{Transformation to object-centric images. It happens when an input image is not looking at the target object. Hence, we transform the image so that it looks at the center of the target object by a homography transformation, which leads to a new pose and a new intrinsic matrix.}
    \label{fig:lookat}
\end{figure}

\textbf{Virtual object-centric camera}. We assume that the object is segmented out for both images. On each image, we denote the 2D object center as $\mathbf{c}$ and the object is bounded by a 2D square bounding box of size $s$, as shown in Fig.~\ref{fig:lookat}. Since the object may not be located at the center of the input image, we need to create a new virtual camera that roughly looks at the object by rotation. In the following, we derive the construction of the intrinsic matrices $\mathbf{K}_{v}^{(1)}$ and $\mathbf{K}_{v}^{(2)}$, the rotation matrices $\mathbf{R}_{v}^{(1)}$ and $\mathbf{R}_{v}^{(2)}$, and corresponding object-centric images $I_{v}^{(1)}$ and $I_{v}^{(2)}$. The virtual cameras are constructed for two input views respectively, so that in the rest of this section, we omit the superscript for simplification.

\textbf{Rotation matrix}. The rotation $\mathbf{R}_v=\mathbf{R}_{v,y}\mathbf{R}_{v,x}$ is computed by rotating around the $y$-axis $\mathbf{R}_{v,y}$ and $x$-axis $\mathbf{R}_{v,x}$ as shown in Fig.~\ref{fig:lookat} to make the camera look at the 2D object center $\mathbf{c}$, which means that 
\begin{equation}
    \mathbf{x}_{v} = \mathbf{R}_{v} \mathbf{x}_{c},
    \label{eq:obj_rot}
\end{equation}
where $\mathbf{x}_v$ is the coordinate of a point in the new virtual camera while $\mathbf{x}_c$ is its coordinate in the input camera.

\textbf{Intrinsic matrix}. Assume that the virtual object-centric camera is constructed to look at the object center $\mathbf{c}$ with a new virtual image size of $s_v$. We determine the new focal length $f_v = s_v \sqrt{f^2+\|\mathbf{c}\|_2^2}  / s$ for this virtual camera, where $f$ is the focal length of the input image. Then, we construct a new intrinsic matrix $\mathbf{K}_v$ with a focal length $f_v$ and a principle point $(s_v/2,s_v/2)$.

\textbf{Object-centric warping}. Since the camera is transformed by a pure rotation, according to multiview geometry~\cite{hartley2003multiple}, the resulting object-centric image and the input image are related by a homography transformation $\mathbf{H}$
\begin{equation}
    \mathbf{H} = \mathbf{K}_v\mathbf{R}_{v}\mathbf{K}^{-1}.
\end{equation}
We use this homography transformation to warp the input image $I$ to get the object-centric image $I_v$. We transform both images into object-centric images $I_{v}^{(1)}$ and $I_{v}^{(2)}$. Therefore, we only need to work on object-centric images and determine their relative transformation $[\mathbf{R}_{v}^{(12)};\mathbf{t}_{v}^{(12)}]$ with 
\begin{equation}\label{eq:obj12}
    \mathbf{x}_{v}^{(1)}= \mathbf{R}_{v}^{(12)}\mathbf{x}_{v}^{(2)}+\mathbf{t}_{v}^{(12)},
\end{equation} 
which can be utilized to determine $[\mathbf{R}^{(12)};\mathbf{t}^{(12)}]$ by Eq.~\ref{eq:obj_rot}.

\subsection{Object Priors from Diffusion Models}
\label{sec:diffusion}
In this section, we aim to utilize the diffusion model Zero123~\cite{zero123} to generate novel-view images $\mathcal{G}=\{I_{g}^{(i)} | i=1,2,3,...,N \}$ of the object. These generated images will be used in the next section to determine the relative pose.




\begin{figure}
    \centering
    \includegraphics[width=0.48\textwidth]{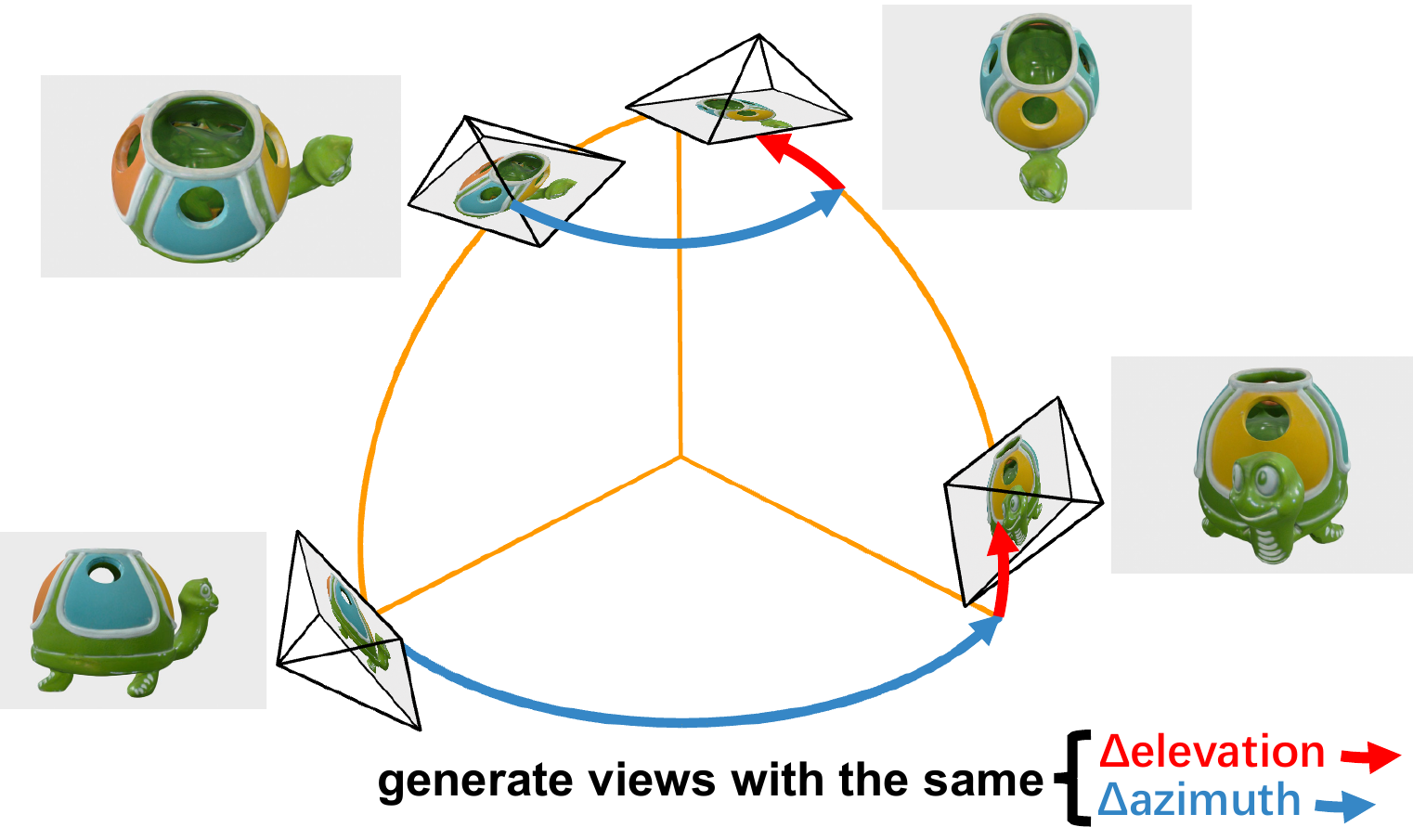}
    \caption{Zero123~\cite{zero123} is able to rotate an given image by a given $\Delta$azimuth and a $\Delta$elevation in the object canonical coordinate. However, Zero123 assumes that the $Y$+ direction (UP) of the image is aligned with gravity direction. Moreover, given the same rotation angle ($\Delta$azimuth and $\Delta$elevation), the actual rotated angle is related to the elevation angle of the input image. This requires us to estimate the inplane rotation and the canonical azimuth of the input image.}
    \label{fig:Zero123IO_and_canonical}
\end{figure}

\textbf{Zero123}~\cite{zero123}. By fine-tuning the large-scale Stable Diffusion~\cite{rombach2022high} model on millions of objects, Zero123~\cite{zero123} learns strong object priors and can generate highly plausible novel-view images from a given image of an object. Given an input image of the object, a delta azimuth, and a delta elevation, Zero123 is able to generate an image on a rotated viewpoint with the given azimuth and elevation changes from the viewpoint of the input view, as shown in Fig.~\ref{fig:Zero123IO_and_canonical}. Zero123 assumes the input image is correctly oriented, which means that the canonical up direction of the object is consistent with the $Y$ direction of the camera coordinate system of the input image. Note that the delta azimuth and delta elevation in Zero123 are also based on the canonical azimuth and elevation of the object. As shown in Fig.~\ref{fig:Zero123IO_and_canonical}, if we apply Zero123 to generate images of the same delta azimuth and delta elevation but on two different images of different canonical elevations, we rotate different angles in the 3D space. We call the coordinate implicitly defined in Zero123 as the canonical object coordinate $\mathbf{x}_o$. In the following, we illustrate how we generate a set of novel-view images using the image $I_v^{(1)}$.

\begin{figure}[t] \centering
\begin{minipage}[t]{0.44\linewidth}
    \includegraphics[width=\textwidth ]{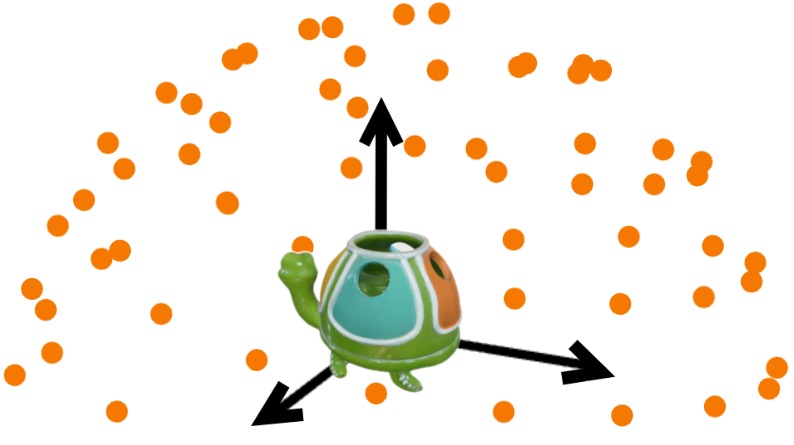}
    \makebox[\textwidth]{\small (a)}
      \label{fig:semiSphere_left}
\end{minipage}
\begin{minipage}[t]{0.52\linewidth}
    \includegraphics[width=\textwidth ]{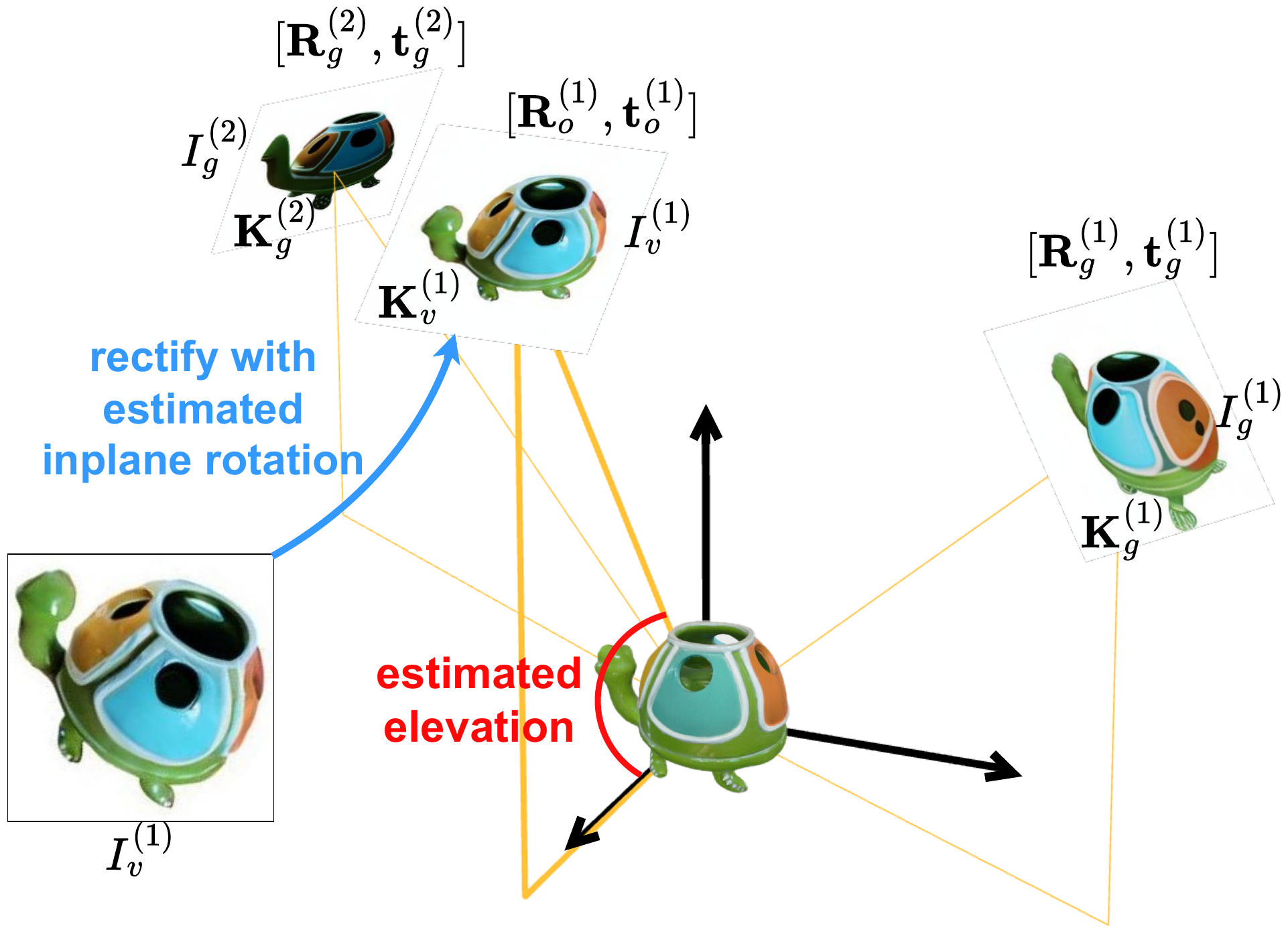}
    \makebox[\textwidth]{\small (b)}
    \label{fig:semiSphere_right}
\end{minipage}\hfill
\caption{(a) We want to generate images on the viewpoints in the canonical object coordinate system. (b) We estimate the inplane rotation and elevation of the image $I^{(1)}_{v}$ and generate images on the given viewpoints using Zero123. For all the generated images and image $I_{v}^{(1)}$, we estimate their new intrinsic matrix and their extrinsic poses in this canonical object coordinate system.} 
\label{fig:semiSphere}
\end{figure}


\textbf{Canonical object viewpoint of $I_v^{(1)}$}. We want to use the Zero123 model to generate a set of novel-view images whose viewpoints are evenly distributed on the upper semi-sphere, as shown in Fig.~\ref{fig:semiSphere} (a). Note that these viewpoints are defined in the canonical object coordinate system with predefined elevations and azimuths. Thus, in order to compute the delta azimuths and delta elevations between $I_v^{(1)}$ and the target novel-view images, we need to compute the canonical azimuth and elevation of image $I_v^{(1)}$.
Meanwhile, we notice that the input image $I_v^{(1)}$ may not be correctly oriented with an in-plane rotation as shown in Fig.~\ref{fig:semiSphere} (b), so we also need to determine an in-plane rotation for $I_v^{(1)}$. Since the azimuth can be arbitrarily defined, we directly set the canonical azimuth of $I_v^{(1)}$ to 0$^\circ$. Thus, we only need to determine the in-plane rotation and the elevation of $I_v^{(1)}$. We extend the elevation determination algorithm of One-2-3-45~\cite{liu2023one} to determine both the elevation and the in-plane rotation. generate a set of hypotheses on the elevations and the in-plane rotations and then select the elevation and the in-plane rotation that can produce the most number of points in the triangulation. More details are included in the Appendix. With the estimated azimuth, elevation, and in-plane rotation, we rectify the input image $I_v^{(1)}$ with the in-plane rotation to be correctly oriented, compute the delta azimuths and elevations, and finally generate predefined novel-view images using Zero123 using the rectified $I_v^{(1)}$.

\textbf{From viewpoints to object poses}. 
In order to utilize the generated images to estimate the relative pose between $I_v^{(1)}$ and $I_v^{(2)}$, we need to explicitly determine the intrinsic matrices and object poses of the generated images and $I_v^{(1)}$ in the canonical object coordinate system. We already derived the intrinsic matrix $K_v^{(1)}$ for $I_v^{(1)}$ and we assume all the generated images have the same intrinsic as $I_v^{(1)}$, i.e. $\mathbf{K}_g^{(i)}=\mathbf{K}_v^{(1)}$. 
To get the object poses $[\mathbf{R}_{g}^{(i)},\mathbf{t}_g^{(i)}]$ and $[\mathbf{R}_o^{(1)},\mathbf{t}_o^{(1)}]$ of the generated images and $I_v^{(1)}$, we already know their elevations and azimuths in the canonical object coordinate system, which helps us to determine the rotations $\mathbf{R}_g^{(i)}$ and $\mathbf{R}_o^{(1)}$. 
Then, assuming that all the generated images and $I_v^{(1)}$ look at the origin of the canonical object coordinate, we only need to determine the distance $d$ to compute the offsets $\mathbf{t}_g^{(i)}$ and $\mathbf{t}_o^{(1)}$. We further assume that the target object is located at the origin inside a sphere with a unit-length radius and the projections of the unit sphere onto every image just inscribe the boundaries of these images, which helps us determine the distance from the origin to the camera center. Thus, with $[\mathbf{R}_{g}^{(i)},\mathbf{t}_g^{(i)}]$ and $[\mathbf{R}_o^{(1)},\mathbf{t}_o^{(1)}]$, we have
\begin{equation}\label{eq:obj1}
    \mathbf{x}_{v}^{(1)} = \mathbf{R}_{o}^{(1)} \mathbf{x}_o + \mathbf{t}_{o}^{(1)},
\end{equation}
\begin{equation}
    \mathbf{x}_{g}^{(i)} = \mathbf{R}_{g}^{(i)} \mathbf{x}_o + \mathbf{t}_{g}^{(i)}.
\end{equation}
where $\mathbf{x}_o$ is the object coordinate and $\mathbf{x}_g^{(i)}$ is the camera coordinate of $i$-th generated image.

\subsection{Object pose estimation}
\label{sec:matching}

In this section, we match the image $I_{v}^{(2)}$ with the posed generated image set $\mathcal{G}=\{(I_g^{(i)}\mathbf{R}_{g}^{(i)},\mathbf{t}_{g}^{(i)}|i=1,...,N\}$ to determine the object pose $[\mathbf{R}_{o}^{(2)},\mathbf{t}_{o}^{(2)}]$ of $I_{v}^{(2)}$ with
\begin{equation}\label{eq:obj2}
    \mathbf{x}_{v}^{(2)} = \mathbf{R}_{o}^{(2)} \mathbf{x}_o + \mathbf{t}_{o}^{(2)}.
\end{equation}
Then, by combining combining Eq.~\ref{eq:obj1} and Eq.~\ref{eq:obj2}, we can get Eq.~\ref{eq:obj12} to compute the relative camera pose between the object-centric images $I_v^{(1)}$ and $I_v^{(2)}$.

To achieve this purpose, we follow the work Gen6D~\cite{liu2022gen6d} to first select a generated image with the most similar viewpoint and then apply a feature volume-based refiner to iteratively refine the final pose. 

\textbf{Viewpoint selection}. We compare $I_v^{(2)}$ with every generated image $I_g^{(i)}$ to compute a matching score. The generated image with a higher score is supposed to be similar to the input image. Similar to \cite{liu2022gen6d}, we also share the information among all generated views by global normalization layers and transformers before computing the final matching scores. We regard the viewpoint of the most similar generated image as the viewpoint of $I_{v}^{(2)}$, which provides a coarse estimation of the object pose $[\mathbf{R}_{o}^{(2)},\mathbf{t}_{o}^{(2)}]$. We then refine the estimated pose with a feature volume-based refiner.

\textbf{Feature volume-based pose refinement}. We use the feature volume-based refiner in Gen6D~\cite{liu2022gen6d}. Given the coarse object pose estimation, we first find the $K$ nearest generated images as reference images for the current refinement step. A 3D feature volume will be constructed using the 2D CNN features extracted from $I_{v}^{(2)}$ and all reference images. Then, a 3D CNN is applied to regress a pose residual in the form of a similarity transformation to update the input coarse estimated pose. This refinement process is iteratively applied for several steps to get an accurate estimation of the object pose $[\mathbf{R}_{o}^{(2)},\mathbf{t}_{o}^{(2)}]$. 

\section{Experiments and Discussions}
\label{sec:experiments}

\begin{figure*}
\begin{center}
    \begin{minipage}{0.99\textwidth}
\begin{center}
\begin{subfigure}{0.85\textwidth}
\includegraphics[width=0.99\linewidth]{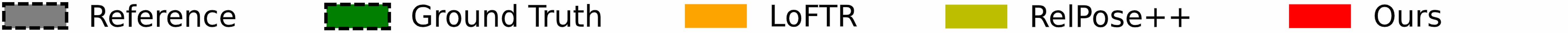}
\end{subfigure}
\end{center}
\vspace{+0.0em}
\end{minipage}
\begin{minipage}{0.9\textwidth}
\begin{center}
\begin{subfigure}{0.3\textwidth}
\includegraphics[width=0.99\linewidth]{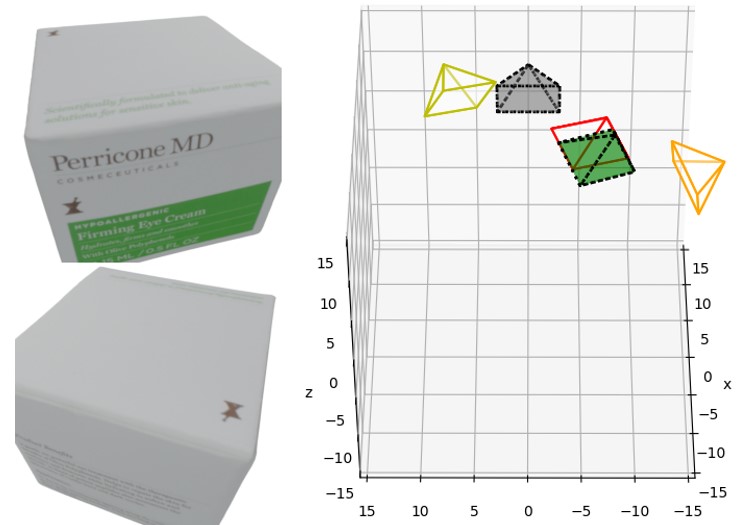}
\end{subfigure}
\begin{subfigure}{0.3\textwidth}
\includegraphics[width=0.99\linewidth]{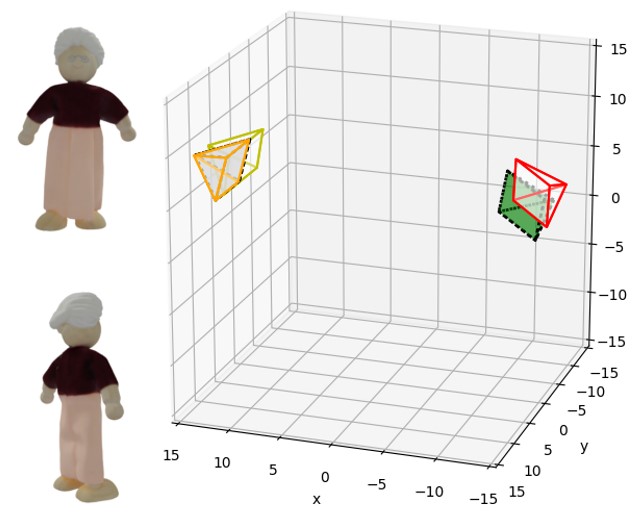}
\end{subfigure}
\begin{subfigure}{0.3\textwidth}
\includegraphics[width=0.99\linewidth]{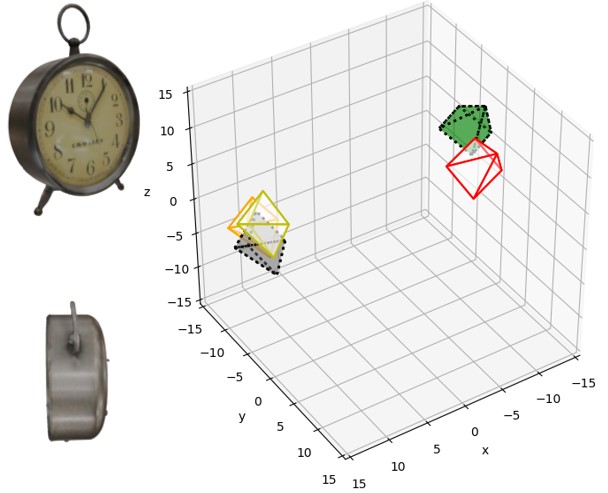}
\end{subfigure}  
\end{center}
\end{minipage}
\begin{minipage}{0.9\textwidth}
\begin{center}
\begin{subfigure}{0.3\textwidth}
\includegraphics[width=0.99\linewidth]{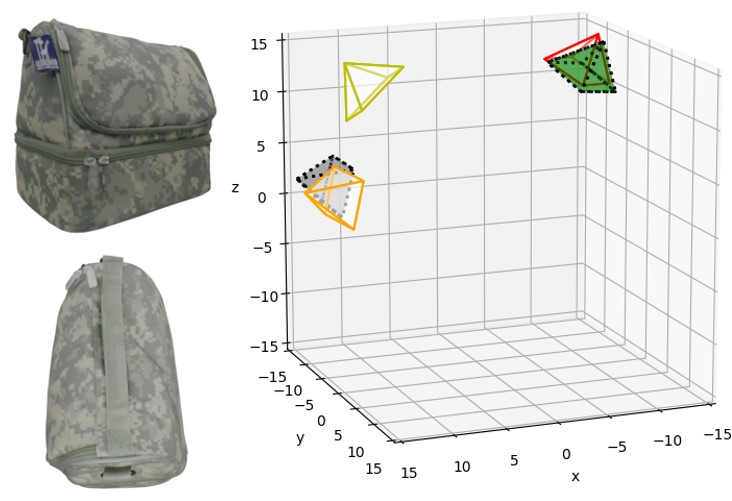}
\end{subfigure}
\begin{subfigure}{0.3\textwidth}
\includegraphics[width=0.99\linewidth]{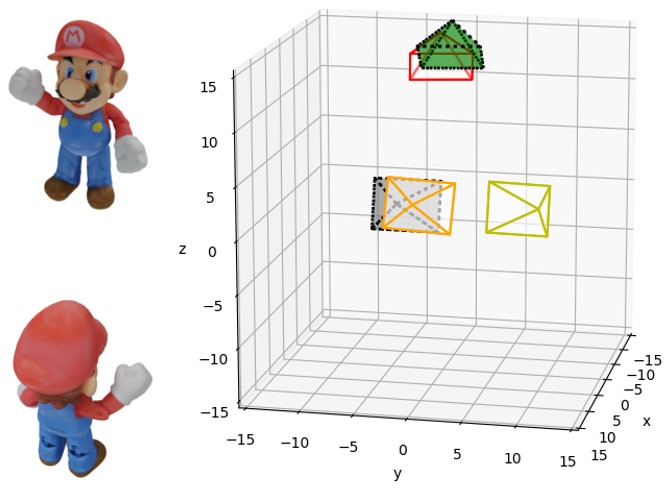}
\end{subfigure}
\begin{subfigure}{0.3\textwidth}
\includegraphics[width=0.99\linewidth]{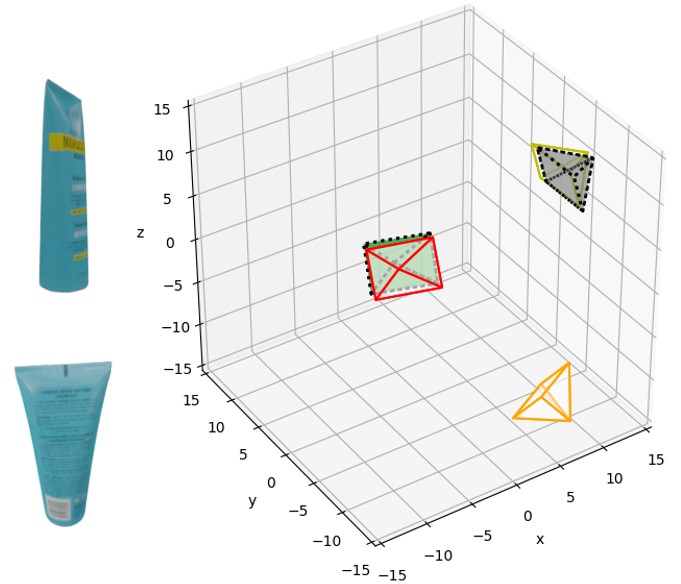}
\end{subfigure}  
\end{center}
\end{minipage}
\end{center}
\caption{Visual Comparison on the GSO dataset~\cite{gso2022google}.}
\label{fig:compGSO}
\end{figure*}

\begin{figure*}
\begin{minipage}{0.99\textwidth}
\begin{center}
\begin{subfigure}{0.85\textwidth}
\includegraphics[width=0.99\linewidth]{images/results/legend_V3.jpg}
\end{subfigure}
\end{center}
\vspace{+0.0em}
\end{minipage}
\begin{minipage}{0.99\textwidth}
\begin{center}
\begin{subfigure}{0.3\textwidth}
\includegraphics[width=0.99\linewidth]{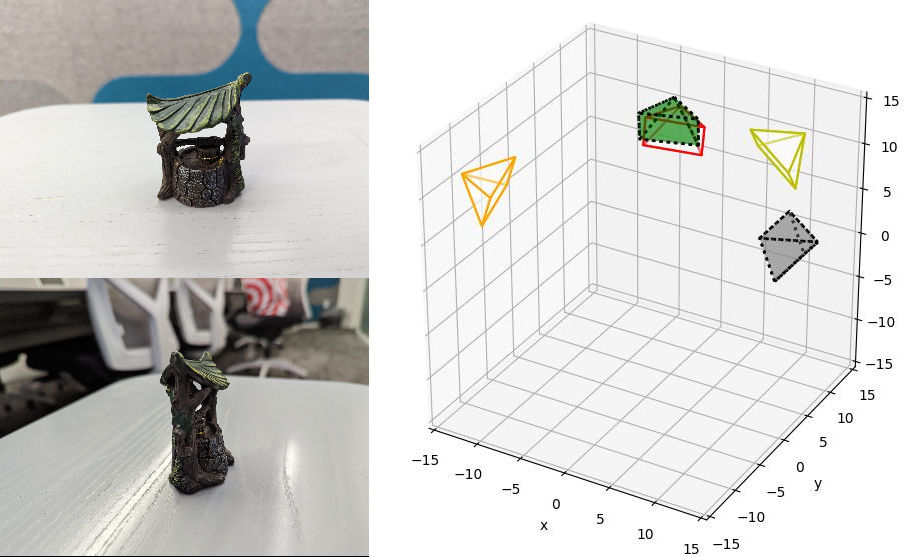}
\end{subfigure}
\begin{subfigure}{0.3\textwidth}
\includegraphics[width=0.99\linewidth]{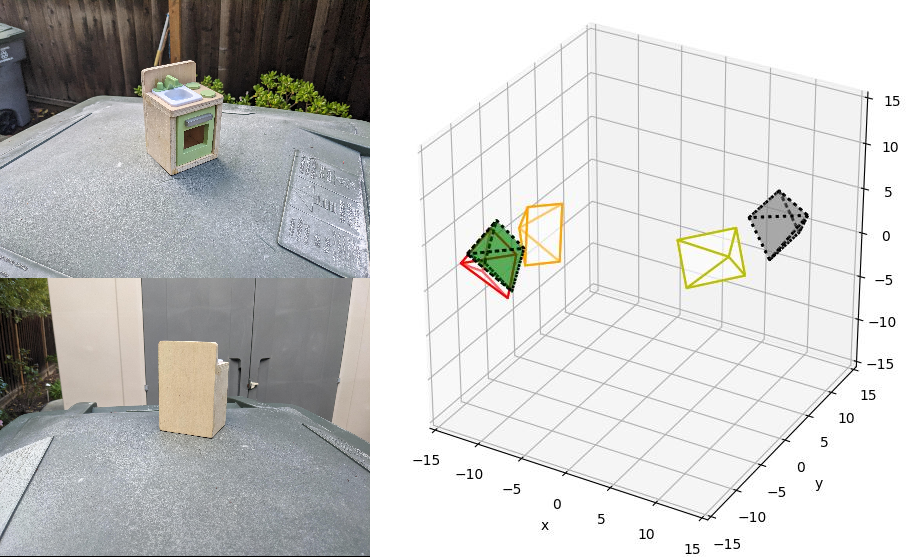}
\end{subfigure}
\begin{subfigure}{0.3\textwidth}
\includegraphics[width=0.99\linewidth]{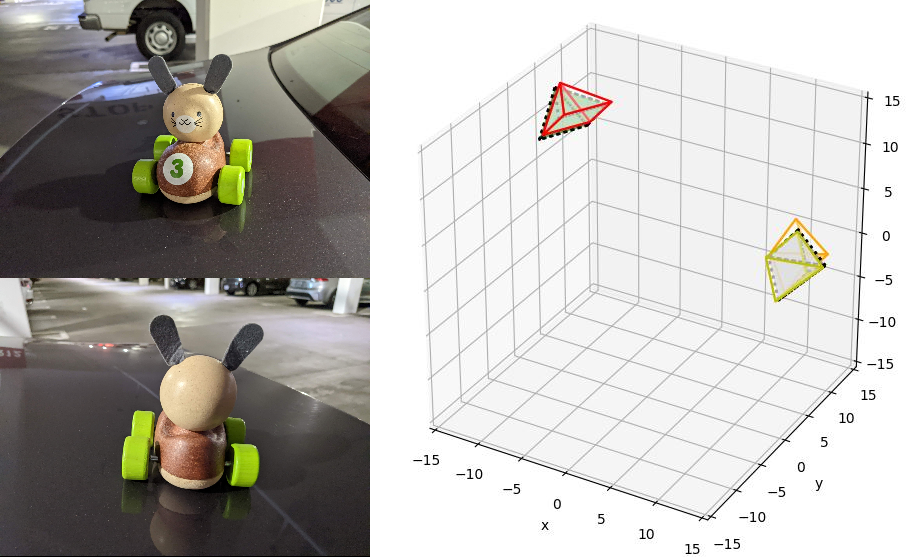}
\end{subfigure}  
\end{center}
\vspace{+0.5em}
\end{minipage}
\begin{minipage}{0.99\textwidth}
\begin{center}
\begin{subfigure}{0.3\textwidth}
\includegraphics[width=0.99\linewidth]{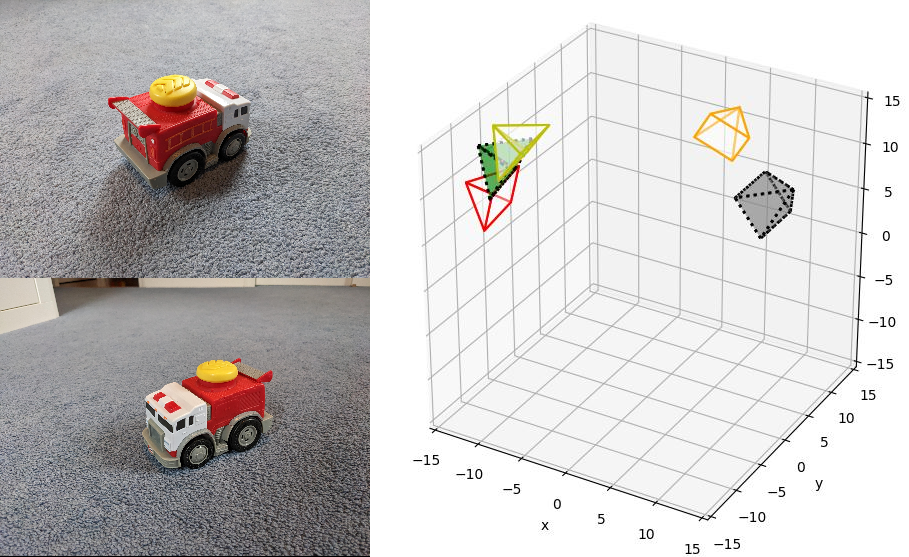}
\end{subfigure}
\begin{subfigure}{0.3\textwidth}
\includegraphics[width=0.99\linewidth]{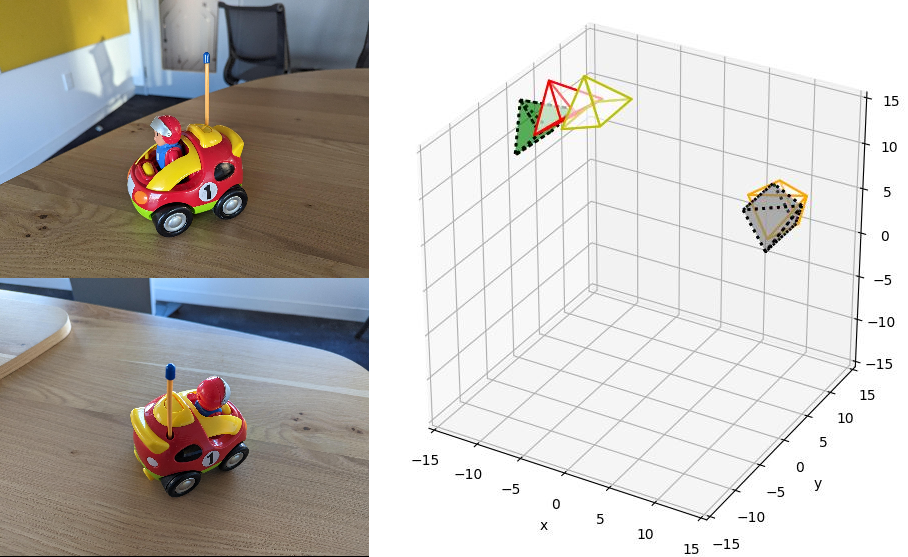}
\end{subfigure}
\begin{subfigure}{0.3\textwidth}
\includegraphics[width=0.99\linewidth]{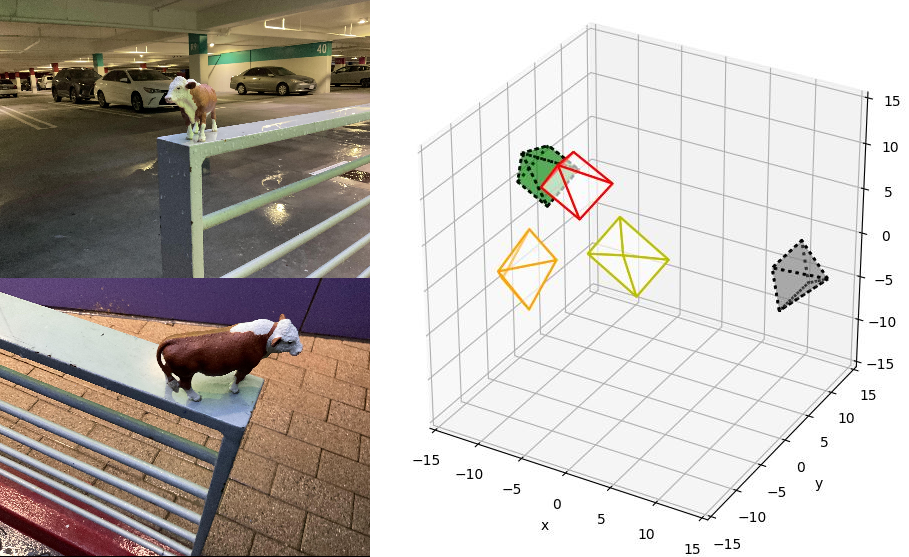}
\end{subfigure}  
\end{center}
\end{minipage}
\caption{Visual Comparison on the Navi dataset~\cite{jampani2023navi} .}
\label{fig:compNavi}
\end{figure*}

\begin{figure*}
\begin{center}
\begin{minipage}{0.9\textwidth}
\vspace{+0em}
\begin{center}
\begin{subfigure}{0.3\textwidth}
\includegraphics[width=0.99\linewidth]{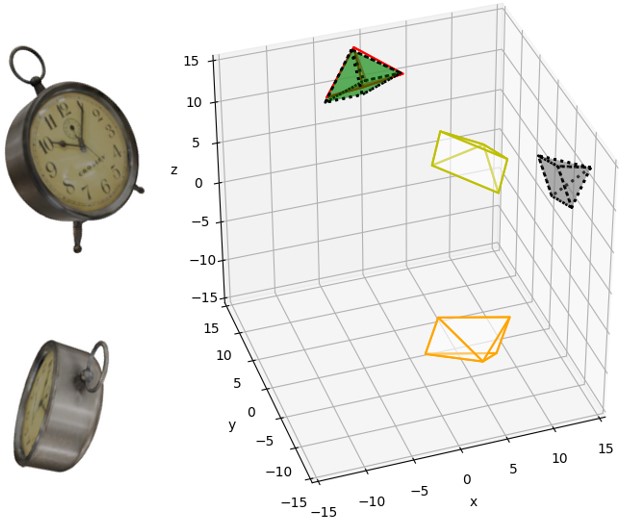}
\end{subfigure}
\begin{subfigure}{0.3\textwidth}
\includegraphics[width=0.99\linewidth]{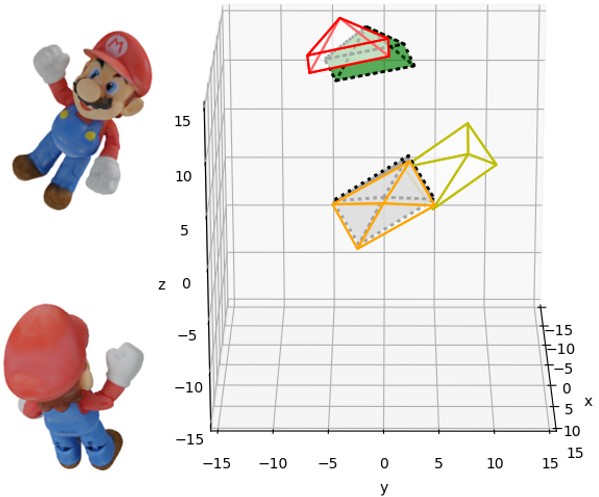}
\end{subfigure}
\begin{subfigure}{0.3\textwidth}
\includegraphics[width=0.99\linewidth]{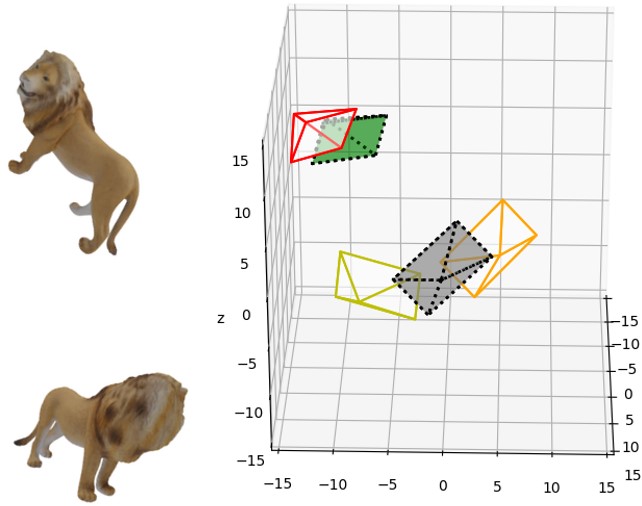}
\end{subfigure}  
\end{center}
\vspace{+0.5em}
\end{minipage}
\begin{minipage}{0.9\textwidth}
\begin{center}
\begin{subfigure}{0.3\textwidth}
\includegraphics[width=0.99\linewidth]{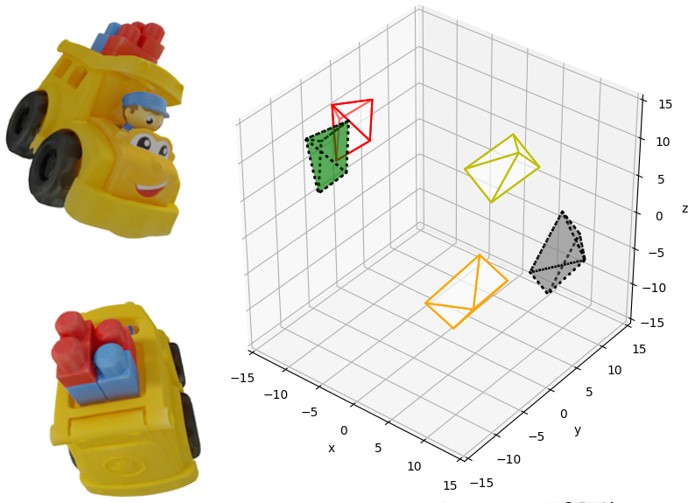}
\end{subfigure}
\begin{subfigure}{0.3\textwidth}
\includegraphics[width=0.99\linewidth]{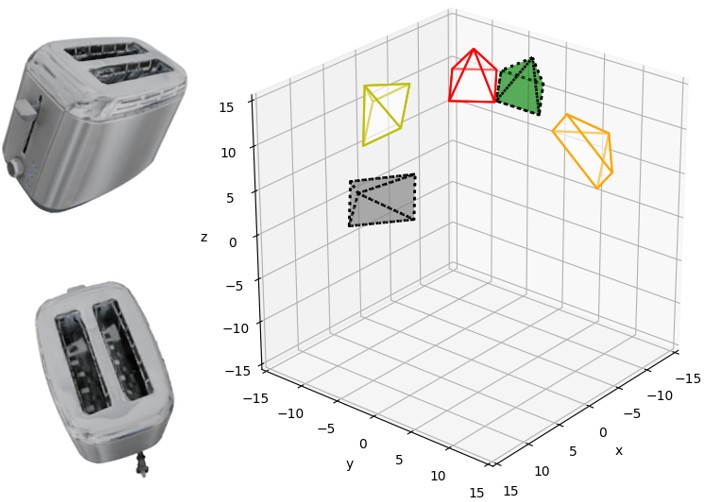}
\end{subfigure}
\begin{subfigure}{0.3\textwidth}
\includegraphics[width=0.99\linewidth]{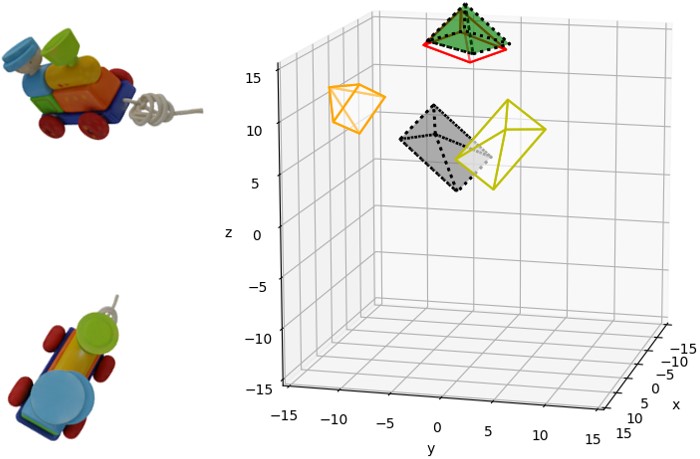}
\end{subfigure}  
\end{center}
\end{minipage}
\end{center}
\caption{Visual Comparison on the rotated GSO dataset~\cite{gso2022google}. }
\label{fig:compRotatedGSO}
\end{figure*}

\subsection{Experimental Setups}

\textbf{Implementation Details} \quad
We utilize Zero123 to generate 128 novel view images $I_g$ for each reference image. When estimating inplane rotation and elevation, the number of diffusion inference steps is 75, and when generating $I_g$, this number is 50. During the inference of a query image, the selector selects from all 128 generated images. In the refinement stage, we apply the refiner iteratively 3 times.

\textbf{Baselines} \quad
To evaluate the effectiveness of the proposed framework, we compare our method with baseline methods including RelPose++~\cite{relpose++2023}, Relative Pose Regression (RPR)~\cite{arnold2022map}, LoFTR~\cite{sun2021loftr}, SIFT~\cite{lowe2004distinctive}-ZoeDepth~\cite{bhat2023zoedepth}-based PnP method and Procrutes method. Both RelPose++~\cite{relpose++2023} and RPR~\cite{arnold2022map} regress the relative poses between images. RelPose++ is a transformer-based model specially designed for objects by training on the CO3D~\cite{co3D2021} dataset. RPR first regresses the 3D-to-3D correspondences from image features and then solves for the pose with a differentiable solver, which is trained end-to-end on large-scale image pair datasets of common scenes. LoFTR~\cite{sun2021loftr} is a dense 2D-to-2D correspondence estimator with a  transformer, on which we solve the relative pose by the RANSAC algorithm. SIFT~\cite{lowe2004distinctive}-ZoeDepth methods first build 2D-to-2D correspondences by SIFT matching and then compute the relative poses by transforming these 2D-to-2D correspondences into 2D-to-3D (PnP-based) or 3D-to-3D (Procrutes-based) correspondences. We adopt the official implementation of RelPose++ and LoFTR with their officially pretrained models. While for RPR and SIFT-ZoeDepth-based method, we adopt the implementation from \cite{arnold2022map}.


\textbf{Metrics} \quad
To evaluate the predicted 6-DoF poses $\{ R_i, t_i\}$ of an object in a target view, we report rotation accuracy and translation accuracy with the standard two-view geometry evaluation metrics, following SuperGlue~\cite{sarlin2020superglue} and OrderAwareNet~\cite{zhang2019learning}.

\subsection{Datasets}
In order to verify the effectiveness and generalization ability of our framework, we evaluate our method and baseline methods on two datasets, the GSO~\cite{gso2022google} dataset and the Navi~\cite{jampani2023navi} dataset. The GSO dataset is a synthetic object dataset containing about 1k 3D-scanned household objects. From the GSO dataset, we select 23 objects and render 21 images on each object for evaluation. The rendered image pairs are set to have extremely large viewpoint changes. The Navi dataset is a real object dataset, which contains images of the same object captured in different environments and viewpoints with accurate calibrations for both intrinsic and extrinsic matrices. The Navi dataset contains 36 objects, among which we select 27 objects for evaluation. To further show our method is more robust to the inplane rotation, we also add randomly inplane rotation between -45 to 45 degrees to each input image. We call the resulting datasets the rotated GSO dataset and the rotated Navi dataset.

\subsection{Metrics}
To evaluate baseline methods and our method, we choose the accuracy under a specific degree as the metrics, which is similar to \cite{sun2021loftr,relpose++2023,relpose2022}. For the rotation accuracy, we compute the relative rotation between the ground-truth rotation $\mathbf{R}_{\text{gt}}$ and the predicted rotation $\mathbf{R}_{\text{pr}}$ by $\mathbf{R}_{\text{gt}}^{\intercal} \mathbf{R}_{\text{pr}}$ and then transform this rotation into the axis-angle form. The resulting rotation angle of this rotation matrix is regarded as the rotation error in angle. The translation contains a scale ambiguity so we compute the translation error as the angle between the normalized ground-truth translation $\mathbf{t}_{\text{gt}}$ and the normalized predicted translation $\mathbf{t}_{\text{pr}}$ by $\arccos{\mathbf{t}_{\text{gt}}^{\intercal}\mathbf{t}_{\text{pr}}}$. We report translation and rotation accuracy under 15$^\circ$ and 30$^\circ$.

\subsection{Comparisons with baseline methods}
\textbf{Comparisons}. We evaluate our method and baselines on two datasets, GSO~\cite{gso2022google} and Navi~\cite{jampani2023navi}. The qualitative results are shown in Fig.~\ref{fig:compGSO} and Fig.~\ref{fig:compNavi}. While quantitative comparison results are provided in Table~\ref{table:compareALL}. As we can see, the matching based method LoFTR~\cite{sun2021loftr} and SIFT~\cite{lowe2004distinctive} does not perform well on both datasets, because the extreme viewpoint change results in very small overlap regions to estimate correspondences. While two regression-based methods RPR~\cite{arnold2022map} and RelPose++~\cite{relpose++2023} show limited generalization ability. Our method, combining the object prior from diffusion models with the object pose estimator, achieves best performance on all metrics, which outperforms baselines by a large margin. More compariosons on the rotated Navi Dateset, as well as visualization of correspondences of LoFTR/SIFT ZoeDepth methods and of depth maps estimated by ZoeDepth, are included in the Appendix.

\textbf{Robustness to inplane rotations}.  We also provided results on the rotated GSO and Navi datasets to show the robustness to inplane rotations. The quantitative results are shown in Table~\ref{table:compareRot} and qualitative results are shown in Fig.~\ref{fig:compRotatedGSO} and Fig.~\ref{fig:compRotatedNAVI}. As we can see, Relpose++~\cite{relpose++2023} show worse robustness to the inplane rotations while our method achieves stronger robustness by estimating an inplane rotation in the pipeline.  




\begin{table*}
\caption{The quantitative comparison results on the two testing datasets.}
\label{table:compareALL}
\centering
\begin{tabular}{lcccccccc}
\toprule
\multirow{3}{*}{Method}                  & \multicolumn{4}{c}{NAVI} & \multicolumn{4}{c}{GSO} \\
 &
  \multicolumn{2}{c}{Rotation  Accuracy} &
  \multicolumn{2}{c}{Translation Accuracy} &
  \multicolumn{2}{c}{Rotation Accuracy} &
  \multicolumn{2}{c}{Translation  Acuracy} \\

 &
  \multicolumn{1}{c}{15$^\circ$} &
  \multicolumn{1}{c}{30$^\circ$} &
  \multicolumn{1}{c}{15$^\circ$} &
  \multicolumn{1}{c}{30$^\circ$} &
  \multicolumn{1}{c}{15$^\circ$} &
  \multicolumn{1}{c}{30$^\circ$} &
  \multicolumn{1}{c}{15$^\circ$} &
  30 \\ 
\midrule
SIFT+ZoeDepth+PnP                        & 19.66 & 25.55 & 12.47 & 25.25       & 7.17 & 13.04 & 5.65 & 15.65      \\
SIFT+ZoeDepth+Procrustes                    & 16.65 & 26.72 & 9.60 & 24.31         & 5.00 & 14.78 & 3.48 & 14.57     \\
map-free-loc RPR   & 17.23 & 33.99 & 13.81 & 35.92           & 4.57 & 15.87 & 3.91 & 16.52   \\
LoFTR                          & 16.59 & 27.99 & 20.74 & 30.38         &  \underline{20.65} & 29.57 &  \underline{29.57} &  \underline{36.52}      \\
Relpose++                        &  \underline{24.33} &  \underline{40.05} &  \underline{24.84} &  \underline{42.71}         & 15.65 &  \underline{32.17} & 20.22 & 32.61     \\
Ours  & \textbf{43.16} & \textbf{66.47} & \textbf{50.64} & \textbf{72.41}       & \textbf{40.43} & \textbf{57.61} & \textbf{42.39} & \textbf{60.65}    \\
\bottomrule
\end{tabular}
\end{table*}


\begin{table*}
\caption{The quantitative comparison results on the two rotated testing datasets.}
\label{table:compareRot}
\centering
\begin{tabular}{lcccccccc}
\toprule
\multirow{3}{*}{Method}                  & \multicolumn{4}{c}{Rotated NAVI} & \multicolumn{4}{c}{Rotated GSO} \\
 &
  \multicolumn{2}{c}{Rotation  Accuracy} &
  \multicolumn{2}{c}{Translation Accuracy} &
  \multicolumn{2}{c}{Rotation Accuracy} &
  \multicolumn{2}{c}{Translation  Acuracy} \\
 &
  \multicolumn{1}{c}{15$^\circ$} &
  \multicolumn{1}{c}{30$^\circ$} &
  \multicolumn{1}{c}{15$^\circ$} &
  \multicolumn{1}{c}{30$^\circ$} &
  \multicolumn{1}{c}{15$^\circ$} &
  \multicolumn{1}{c}{30$^\circ$} &
  \multicolumn{1}{c}{15$^\circ$} &
  30 \\ 
\midrule
SIFT+ZoeDepth+PnP            & 18.09 & 23.90 & 13.68 & 22.08          & 7.83 & 14.35 & 5.65 & 17.61\\
SIFT+ZoeDepth+Procrustes    & 15.47 & 23.95 & 8.11 & 19.31       & 5.00 & 13.70 & 2.83 & 14.78      \\
Map-free-loc RPR            & 8.69 & 17.58 & 7.78 & 22.23     & 2.61 & 9.78 & 5.87 & 18.91     \\
LoFTR                       & \underline{13.14} & 21.28             & \underline{17.38}       &28.97          &  \underline{14.78}  &  \underline{18.70} &  \underline{16.52} &  \underline{24.35}    \\
Relpose++                   & 12.10             & \underline{26.46} & 12.74 &  \underline{29.20}   & 6.52               & 16.96 & 6.74 & 20.00     \\
Ours  & \textbf{29.81} & \textbf{49.34} & \textbf{35.36} & \textbf{60.27}           & \textbf{31.30} & \textbf{48.04} & \textbf{36.30} & \textbf{56.74}     \\
\bottomrule
\end{tabular}
\end{table*}




\subsection{Analysis}
In this section, we conduct analysis on each module proposed in this paper to demonstrate their effectiveness. 

\textbf{Inplane rotation and elevation prediction}. As stated in Sec.~\ref{sec:diffusion}, we need to estimate an inplane rotation and an elevation for the input image $I_{v}^{(1)}$ so that we can correctly generate images on the desired viewpoints by Zero123~\cite{zero123}. To show this necessity, we directly treat the inplane rotation as 0$^{\circ}$ and elevation as 45$^{\circ}$, and evaluate the performances on the rotated datasets. The results are listed in Table~\ref{table:ablationRot}. As the results shown, it is vital to estimate both inplane rotation and elevation to get a correct relative pose estimation in our pipeline.

\begin{table*}
\caption{Ablation studies on inplane rotations and elevation predictions.}
\label{table:ablationRot}
\centering
\begin{tabular}{lcccccccc}
\toprule
\multirow{3}{*}{Method}                  & \multicolumn{4}{c}{Rotated NAVI} & \multicolumn{4}{c}{Rotated GSO} \\
 &
  \multicolumn{2}{c}{Rotation  Accuracy} &
  \multicolumn{2}{c}{Translation Accuracy} &
  \multicolumn{2}{c}{Rotation Accuracy} &
  \multicolumn{2}{c}{Translation  Acuracy} \\
 &
  \multicolumn{1}{c}{15$^\circ$} &
  \multicolumn{1}{c}{30$^\circ$} &
  \multicolumn{1}{c}{15$^\circ$} &
  \multicolumn{1}{c}{30$^\circ$} &
  \multicolumn{1}{c}{15$^\circ$} &
  \multicolumn{1}{c}{30$^\circ$} &
  \multicolumn{1}{c}{15$^\circ$} &
  30 \\ 
\midrule
Without predict inplane rotation   & 22.26 & 41.60 & 26.89 & 51.80       & 19.13 & 32.17 & 19.57 & 41.09      \\
Without predict elevation  & 22.34 & 44.33 & 27.34 & 55.87             & 18.91 & 36.09 & 21.74 & 50.87     \\
Without both      & 20.57 & 40.12 & 26.41 & 52.85          & 11.52 & 29.35 & 18.48 & 37.83      \\
With both    & 29.81 & 49.34 & 35.36 & 60.27           & 31.30 & 48.04 & 36.30 & 56.74    \\
\bottomrule
\end{tabular}
\end{table*}

\begin{table*}
\caption{Ablation studies on the number of generated images.}
\label{table:generate}
\centering
\begin{tabular}{ccccccccc}
\toprule
\multirow{3}{*}{The number of generated images} & \multicolumn{4}{c}{NAVI} & \multicolumn{4}{c}{GSO} \\
 &
  \multicolumn{2}{c}{Rotation  Accuracy} &
  \multicolumn{2}{c}{Translation Accuracy} &
  \multicolumn{2}{c}{Rotation Accuracy} &
  \multicolumn{2}{c}{Translation  Acuracy} \\
 &
  \multicolumn{1}{c}{15$^\circ$} &
  \multicolumn{1}{c}{30$^\circ$} &
  \multicolumn{1}{c}{15$^\circ$} &
  \multicolumn{1}{c}{30$^\circ$} &
  \multicolumn{1}{c}{15$^\circ$} &
  \multicolumn{1}{c}{30$^\circ$} &
  \multicolumn{1}{c}{15$^\circ$} &
  30 \\ 
\midrule
8   & 16.98 & 40.54 & 23.69 & 47.45      & 16.52 & 38.04 & 23.91 & 45.43      \\
16  & 29.92 & 54.32 & 36.18 & 62.07          & 31.74 & 51.52 & 36.30 & 58.70      \\
32    & 42.09 & 62.81 & 46.16 & 68.69     & 37.39 & 54.57 & 39.13 & 59.13   \\
64    & 43.10 & 63.93 & 48.19 & 71.61    & 41.09 & 55.22 & 42.83 & 60.65  \\
128    & 43.16 & 66.47 & 50.64 & 72.41      & 40.43 & 57.61 & 42.39 & 60.65      \\
\bottomrule
\end{tabular}
\end{table*}

\begin{table}
\caption{Ablation studies on the number of refinement iterations on the GSO datasets.}
\label{table:refine}
\centering
\begin{tabular}{lcccccccc}
\toprule
\multirow{3}{*}{\#Refine}   & \multicolumn{4}{c}{GSO} \\
 &
  \multicolumn{2}{c}{Rotation  Accuracy} &
  \multicolumn{2}{c}{Translation  Acuracy} \\
 &
  \multicolumn{1}{c}{15$^\circ$} &
  \multicolumn{1}{c}{30$^\circ$} &
  \multicolumn{1}{c}{15$^\circ$} &
  30 \\ 
\midrule
0 & 26.96 & 52.39 & 31.74 & 57.17 \\
1& 36.30 & 55.00 & 38.04 & 60.00 \\
2& 40.22 & 56.74 & 42.39 & 60.22  \\
3& 40.43 & 57.61 & 42.39 & 60.87 \\
4& 41.09 & 57.83 & 42.39 & 60.65 \\
5 & 41.52 & 57.17 & 42.39 & 61.30  \\
6& 40.65 & 57.39 & 43.26 & 61.74  \\
7& 40.87 & 56.96 & 43.04 & 61.30 \\
8& 40.43 & 57.17 & 43.70 & 61.74  \\
\bottomrule
\end{tabular}
\end{table}

\begin{table}
\caption{Ablation studies on sensitiveness to input masks on the NAVI dataset. We use different dilation kernels to dilate the input masks and evaluate the results on inaccurate masks.}
\label{table:mask}
\centering
\begin{tabular}{lcccccccc}
\toprule
\multirow{3}{*}{Dilation}   & \multicolumn{4}{c}{NAVI} \\
 &
  \multicolumn{2}{c}{Rotation  Accuracy} &
  \multicolumn{2}{c}{Translation  Acuracy} \\
 &
  \multicolumn{1}{c}{15$^\circ$} &
  \multicolumn{1}{c}{30$^\circ$} &
  \multicolumn{1}{c}{15$^\circ$} &
  30$^\circ$ \\ 
\midrule
10\%   & 36.38 & 56.43 & 36.18 & 64.31 \\
5\% & 33.06 & 56.11 & 38.69 & 67.37 \\
2\% & 42.23 & 61.70 & 44.61 & 72.65 \\
0\%  & 43.16 & 66.47 & 50.64 & 72.41  \\
\bottomrule
\end{tabular}
\end{table}

\textbf{The number of generated images}. By default, we generate 128 images with Zero123 to estimate the relative pose. To show how our method performs with different numbers of generated images, we conduct an experiment on both test datasets as shown in Table~\ref{table:generate}. As we can see, the performance is almost the same with only 64 or 32 images but drops reasonably with only 8 or 16 images due to the sparsity.

\textbf{The number of refinement iterations}. We further study how the number of iterations in the pose refinement in Sec.~\ref{sec:matching} affects the quality of two-view pose estimation. We show the results on the GSO dataset with refinement number ranging from 0 to 8. As we can see, the performance increases with 0 to 3 iterations but does not further increase with more refinement steps.

\textbf{Sensitiveness to input masks}. To show that our method is robust to the input mask of the object, we manually add dilations to the input mask by 0\% to 10\%, and then evaluate the results on the NAVI dataset. As we can see, adding 2\% dilation to the masks almost does not change the performance at all,  while further adding 10\% dilation will decrease the performance reasonably. Note that with recent SAM~\cite{kirillov2023segment} could produce very accurate object masks on arbitrary, so that the errors of mask are not very large in most cases.

\subsection{Application in Visual Odometry}
\begin{figure}
    \centering
    \includegraphics[width=0.48\textwidth]{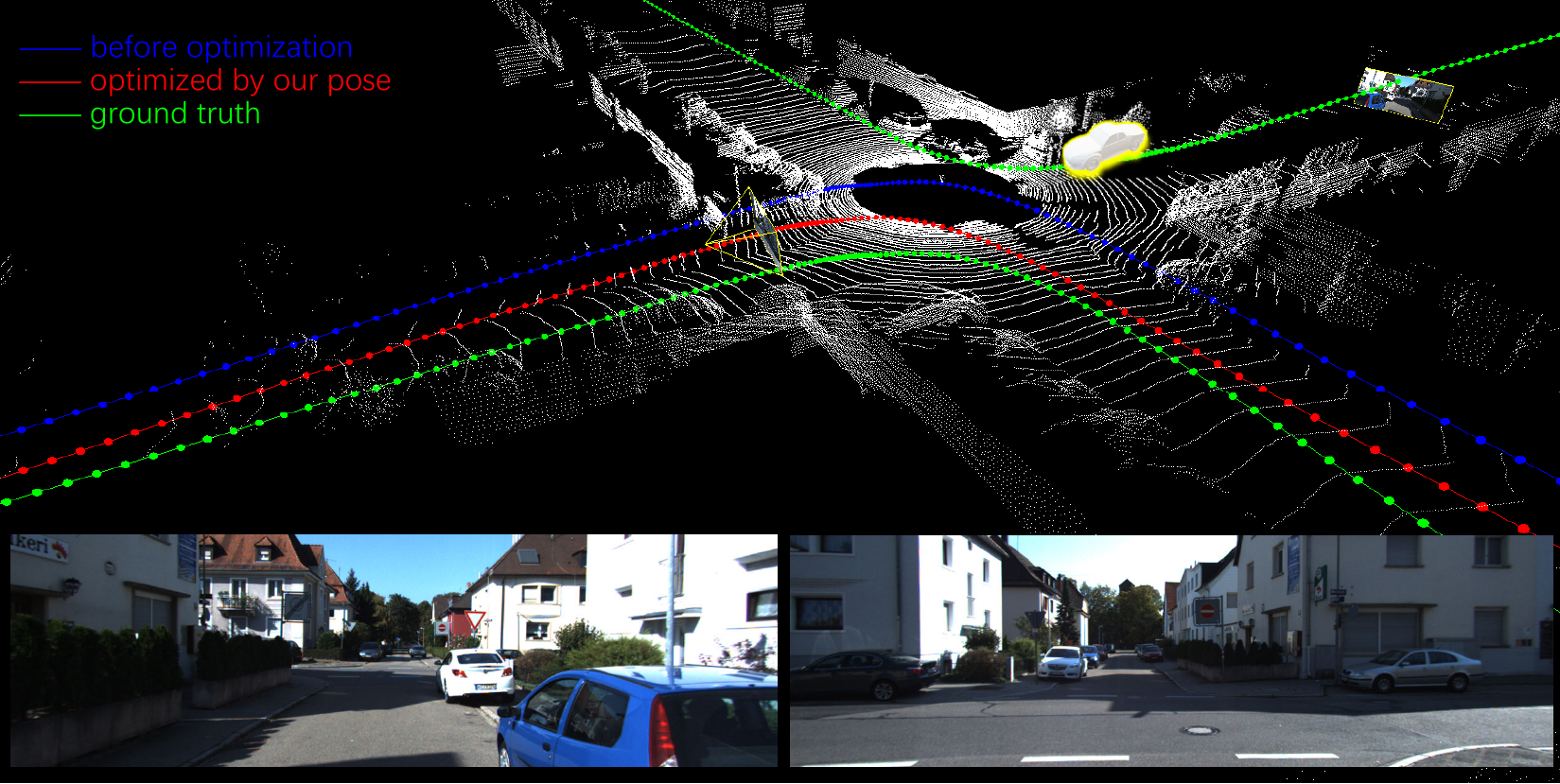}
    \caption{We add an edge to the pose graph based on estimated relative pose from reference image (left) to query image (right), and subsequently employ g2o to perform pose graph optimization. }
    \label{fig:application_in_VO}
\end{figure}

We show an application of our relative pose estimation in the visual odometry (VO) task. We use the ORB-SLAM2~\cite{campos2021orb} track a car to travel around a street and revisit a crossroad with a parking car beside the road as shown in Fig.~\ref{fig:application_in_VO}. In this case, we use our algorithm to estimate the relative pose between two views that are co-visible to this car and then we add the estimated relative pose to the pose graph to optimize it. It can be observed that with our relative pose estimation, the tracked cameras are more accurate and more close to the ground-truth.

\subsection{Runtime Analysis}
All experiments are conducted on an Intel(R) Xeon(R) Gold 5220R CPU @ 2.20GHz and a single NVIDIA GeoForce RTX 3090 GPU. For the configurations, we used in the experiment, when the Zero123 model has already been loaded to GPU, building from a reference image costs eight minutes. After that, for each query image, it takes 0.18s to estimate the pose relative to the reference image. For a faster configuration with a rotation accuracy at 30-degree decrease of only 0.01 on rotated testing dataset, it takes 125s to build from a reference image, including 58s for inplane rotation and elevation prediction, and 67s for novel-view image generations. If the object in reference image is correctly oriented, we can skip the inplane rotation prediction, then the building process only takes 73s under such a faster configuration.

\subsection{Limitations}

\begin{figure}[t]
    \centering
    \includegraphics[width=0.4\textwidth]{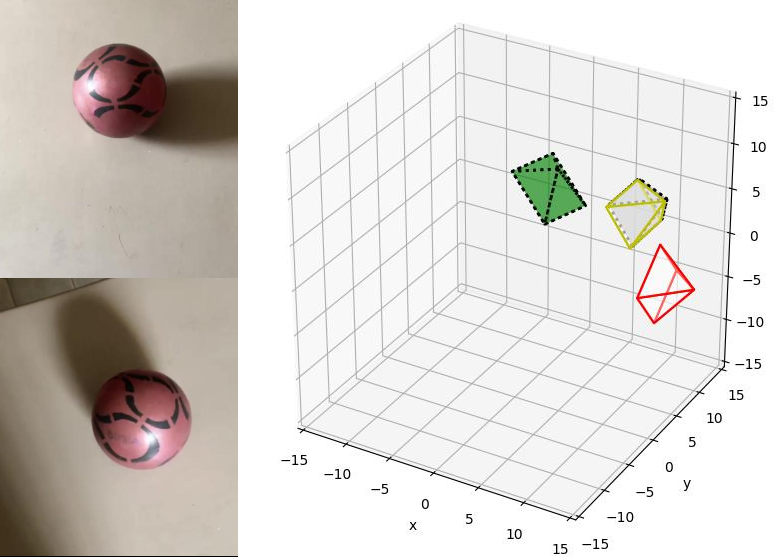}
    \caption{A failure example on a symmetric object.}
    \label{fig:fail_on_symmetric_co3d_obj}
\end{figure}

A failure example of our method on a symmetric object is shown in Fig.~\ref{fig:fail_on_symmetric_co3d_obj}. 
However, precisely predicting the poses of symmetrical objects is really an ill-posed problem in the extreme two-views setting.  
Although Relpose++~\cite{relpose++2023} attempted to handle object symmetries, they rely on additional views (more than 2 views) to avoid ambiguity. RelPose++ also fails in the symmetric example as shown in Fig.~\ref{fig:fail_on_symmetric_co3d_obj}

\section{Conclusion}
In this paper, we introduced a new algorithm to estimate relative camera poses with extreme viewpoint changes. The key idea of our method is to utilize the object prior learned from large-scale 2D diffusion model Zero123~\cite{zero123}, which is able to generate novel-view images of an object. However, since Zero123 has an canonical coordinate system implicitly defined in its model and the image may not look at the object, we cannot directly apply Zero123. To address this challenge, we first propose a new formulation of the two-view pose estimation as an object pose estimation problem and correctly define the object poses for both input images and the generated images. Finally, we match the other image against the generated images to get an object pose estimation, which helps us determine the relative two-view camera poses. Extensive experiments on the GSO and the NAVI dataset have demonstrate the effectiveness of our design. We have also show an application of our method in a Visual Odometry system.

\bibliographystyle{IEEEtran}
\bibliography{sparsePose}


\appendix
\subsection{Implementation Details on Inplane Rotation and Elevation Predictor}
When the object is correctly oriented, zero123 is expected to generate images with the highest consistency. We enumerate all possible inplane rotation angles, and for each inplane rotation candidate we rotate the image accordingly and assess the consistency under this rotated image. The inplane rotation angle associated with the highest consistency is considered as the predicted inplane rotation angle. 

To measure the consistency under an input image, we follow the elevation predict module in One2345\cite{liu2023one} to firstly generate several nearby views of this input image, and then calculate the re-projection error for each possible elevation angle. The smallest re-projection error is considered as the re-projection error under this input image. 

To accelerate the enumeration process, similar to the elevation prediction module in One2345 \cite{liu2023one}, we employ a coarse-to-fine strategy consisting of three stages. For instance, if the maximum inplane rotation angle of the image is ±45$^\circ$, in the first stage we enumerate ($-30^\circ$, $-10^\circ$, $10^\circ$, $30^\circ$). In the subsequent stage, the interval decreases to 7$^\circ$, meaning we enumerate ($x-14^\circ$, $x-7^\circ$, $x+7^\circ$,$x+14^\circ$), where $x$ represents the angle with the highest consistency in the first stage; Finally, in the last stage the interval is further reduced to $2^\circ$.

\subsection{More Experimental Results}

\begin{figure*}[t]
\begin{minipage}{0.99\textwidth}
\begin{center}
\begin{subfigure}{0.85\textwidth}
\includegraphics[width=0.99\linewidth]{images/results/legend_V3.jpg}
\end{subfigure}
\end{center}
\vspace{+0.0em}
\end{minipage}
\begin{minipage}{0.95\textwidth}
\begin{center}
\begin{subfigure}{0.3\textwidth}
\includegraphics[width=0.99\linewidth]{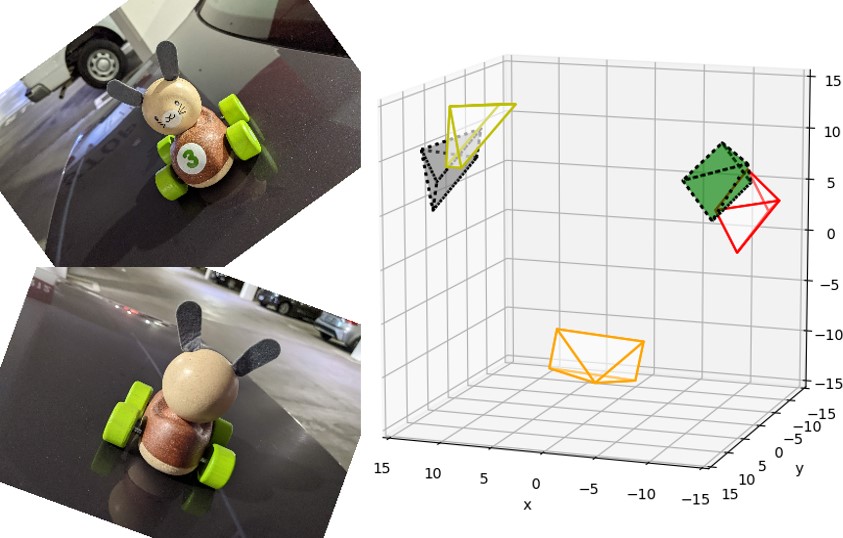}
\end{subfigure}
\begin{subfigure}{0.3\textwidth}
\includegraphics[width=0.99\linewidth]{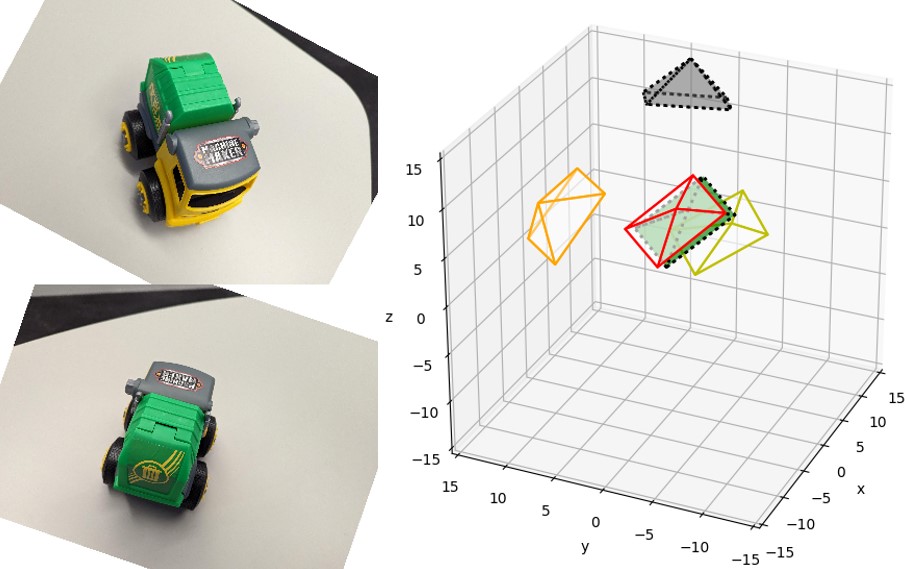}
\end{subfigure}
\begin{subfigure}{0.3\textwidth}
\includegraphics[width=0.95\linewidth]{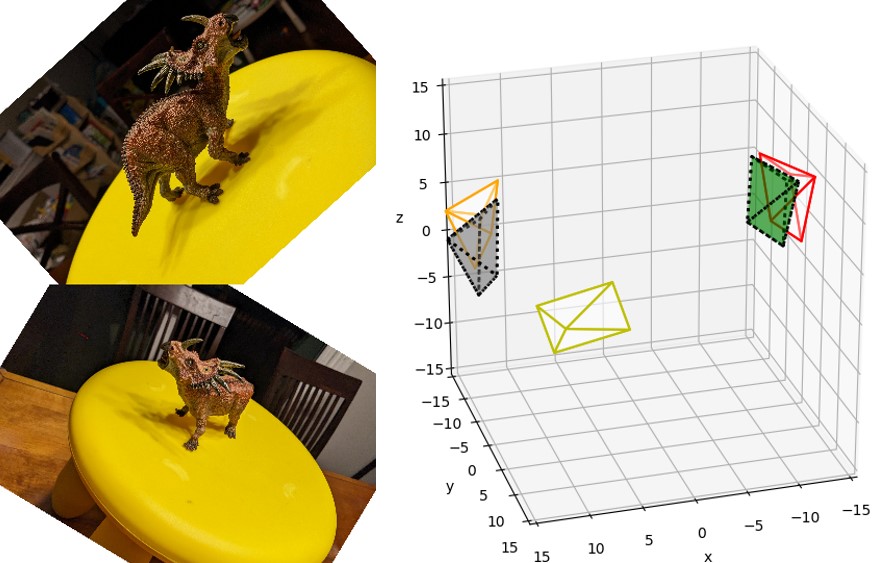}
\end{subfigure}  
\end{center}
\vspace{+0.5em}
\end{minipage}
\begin{minipage}{0.95\textwidth}
\begin{center}
\begin{subfigure}{0.3\textwidth}
\includegraphics[width=0.99\linewidth]{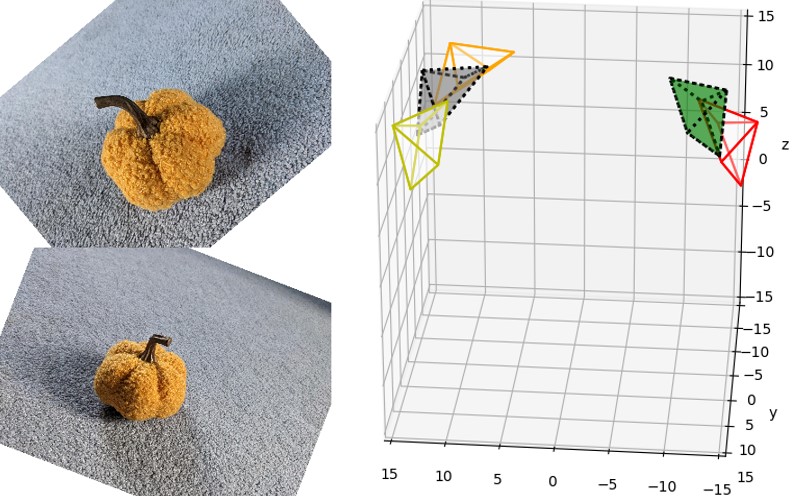}
\end{subfigure}
\begin{subfigure}{0.3\textwidth}
\includegraphics[width=0.99\linewidth]{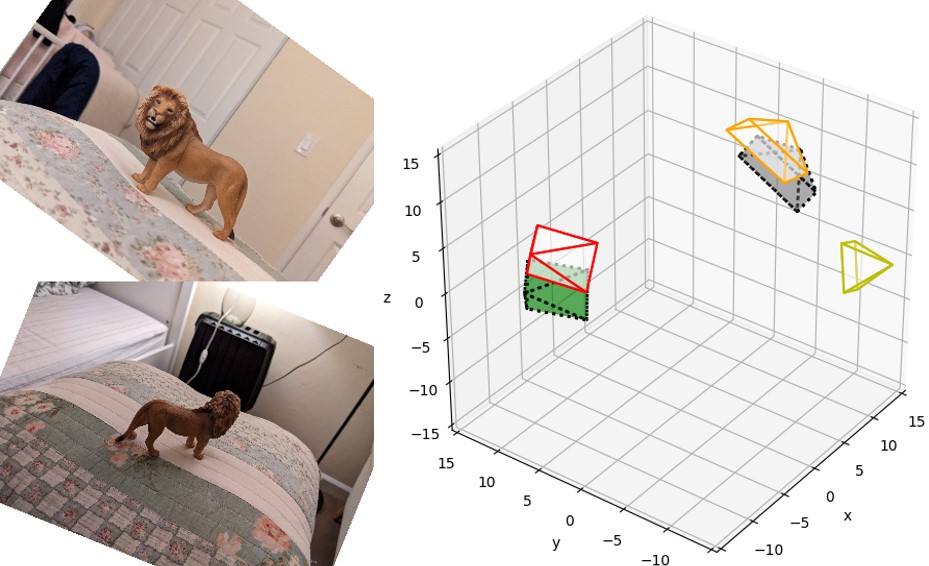}
\end{subfigure}
\begin{subfigure}{0.3\textwidth}
\includegraphics[width=0.99\linewidth]{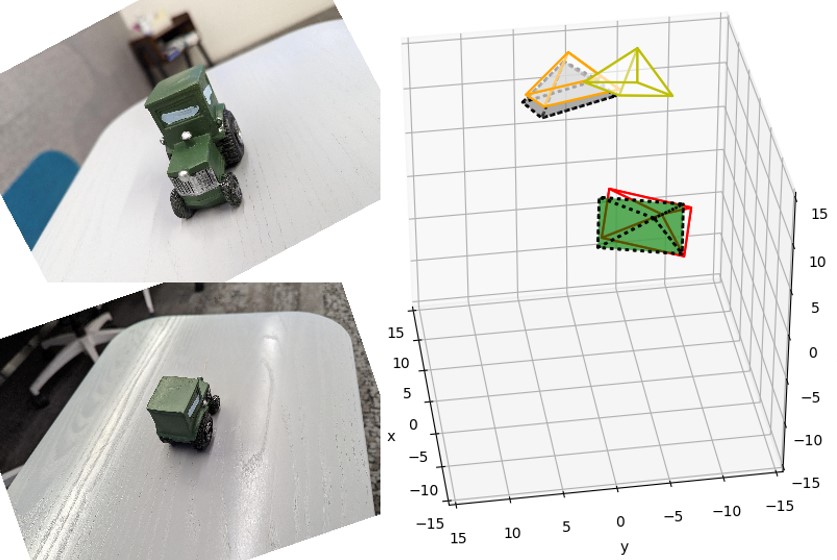}
\end{subfigure}  
\end{center}
\end{minipage}
\caption{Visual comparisons on the rotated NAVI dataset~\cite{jampani2023navi} }
\label{fig:compRotatedNAVI}
\end{figure*}



\subsubsection{Visual results on the rotated NAVI dataset}
\label{sec:suppRotatedNAVI}
Please refer to Fig.~\ref{fig:compRotatedNAVI} for the visual comparisons with the baselines on the rotated NAVI dataset.



\subsubsection{Visualization of correspondences of LoFTR/SIFT-ZoeDepth methods}
\begin{figure}[!htb]
    \centering
    \includegraphics[width=0.4\textwidth]{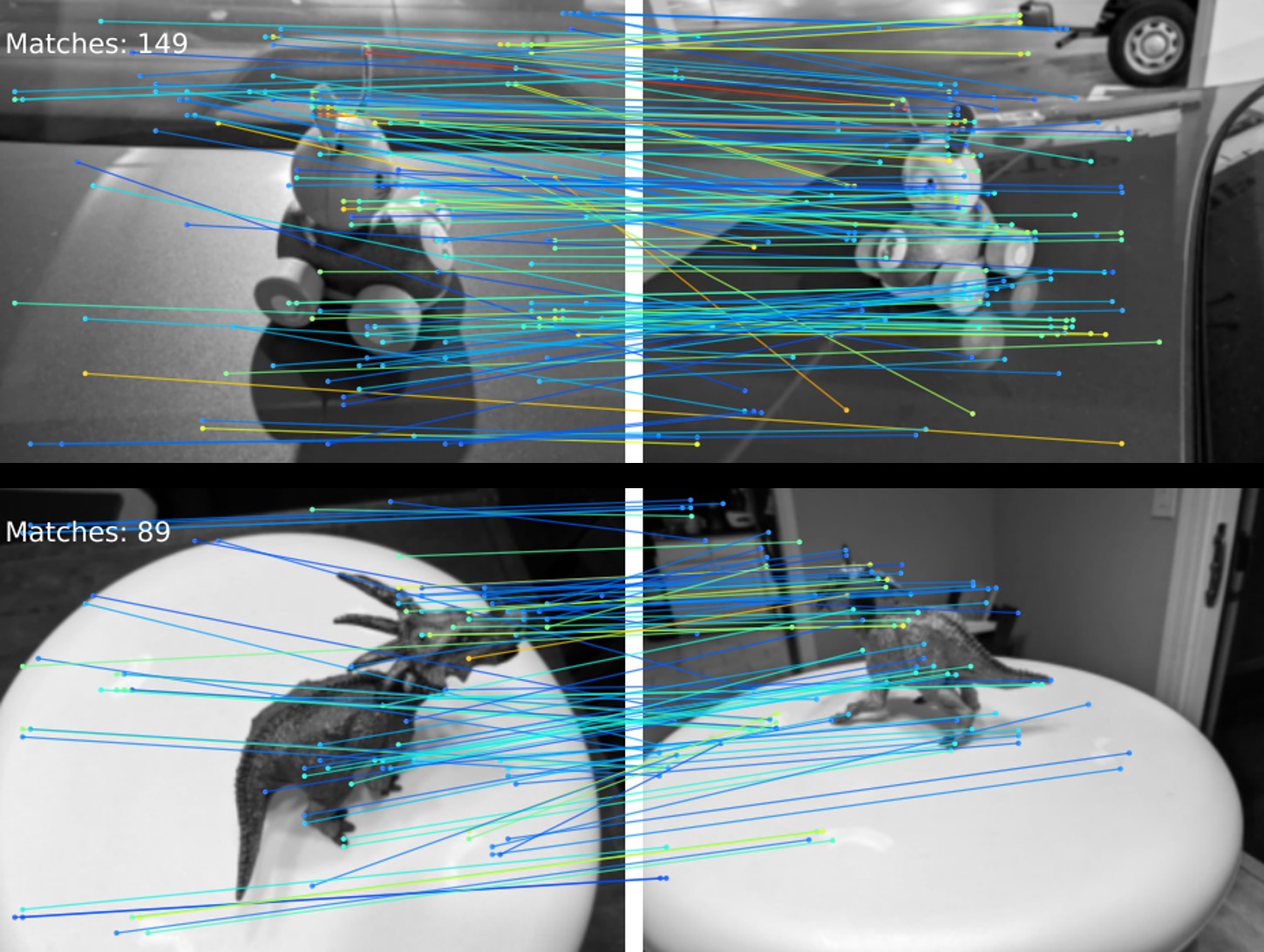}
    \caption{Correspondences of LoFTR on our test set}
    \label{fig:loftr_correspndence_V2}
\end{figure}

Please refer to Fig.~\ref{fig:loftr_correspndence_V2} for the visualization of correspondences of LoFTR on our test set.

\subsubsection{Visualization of depth maps estimated by ZoeDepth}
\begin{figure}[!htb] \centering
    \makebox[0.01\textwidth]{}
    \makebox[0.15\textwidth]{\scriptsize fire engine toy}
    \makebox[0.15\textwidth]{\scriptsize bull}
    \makebox[0.15\textwidth]{\scriptsize dinosaur}
    \\
    \raisebox{0.1\height}{\makebox[0.01\textwidth]{\rotatebox{90}{\makecell{\scriptsize raw}}}}
    \includegraphics[width=0.14\textwidth]{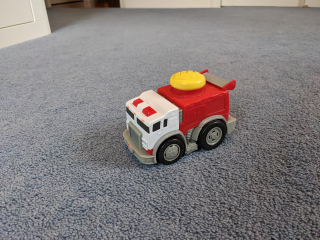}
    \includegraphics[width=0.14\textwidth]{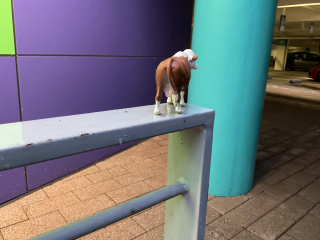}
    \includegraphics[width=0.14\textwidth]{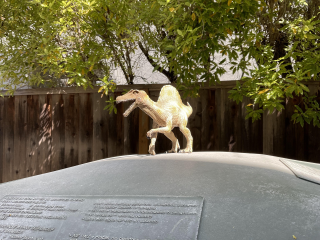}
    \\
    \raisebox{0.1\height}{\makebox[0.01\textwidth]{\rotatebox{90}{\makecell{\scriptsize depth}}}}
    \includegraphics[width=0.14\textwidth]{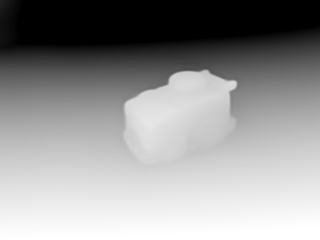}
    \includegraphics[width=0.14\textwidth]{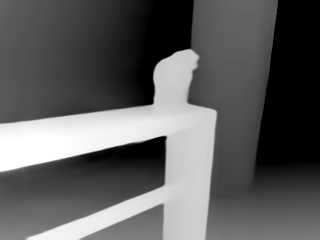}
    \includegraphics[width=0.14\textwidth]{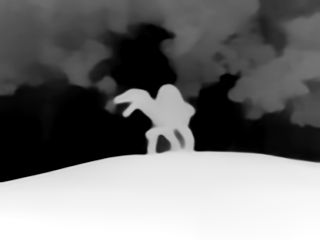}
    \\
    \raisebox{0.1\height}{\makebox[0.01\textwidth]{\rotatebox{90}{\makecell{\scriptsize 3D points}}}}
    \includegraphics[width=0.14\textwidth]{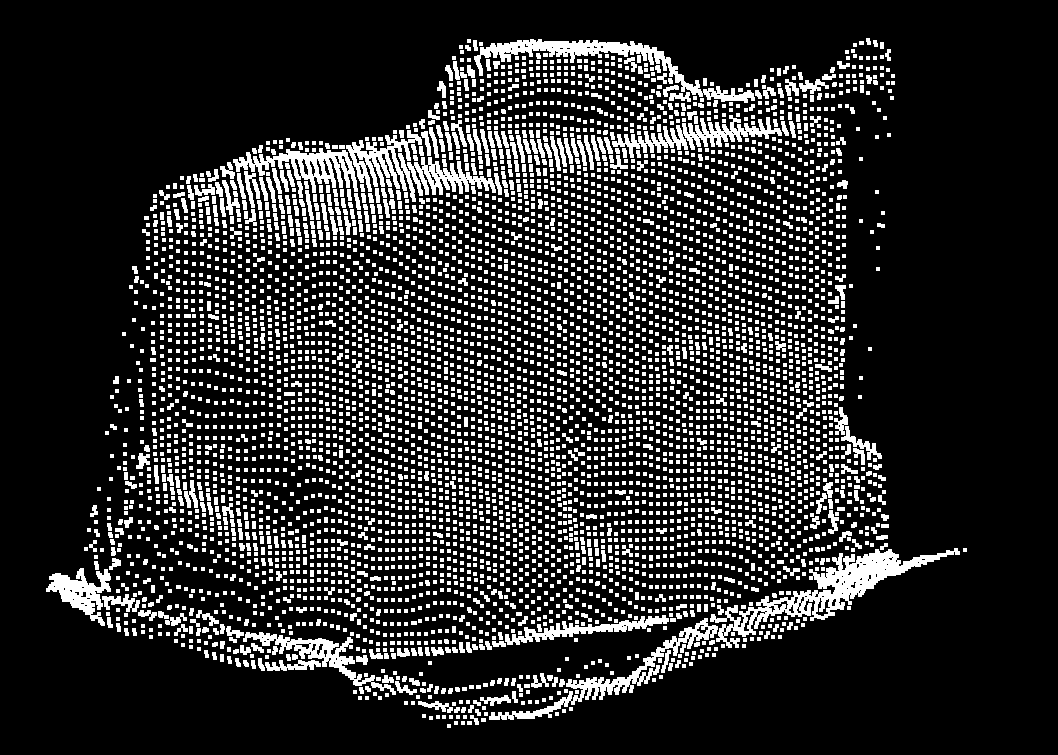}
    \includegraphics[width=0.14\textwidth]{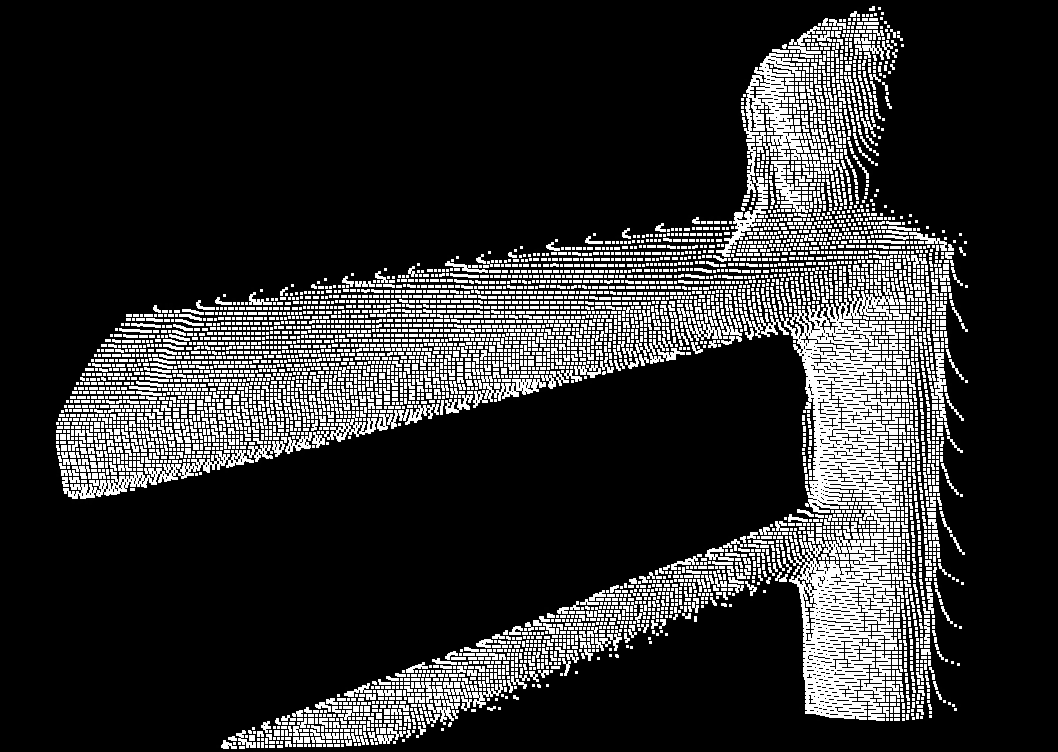}
    \includegraphics[width=0.14\textwidth]{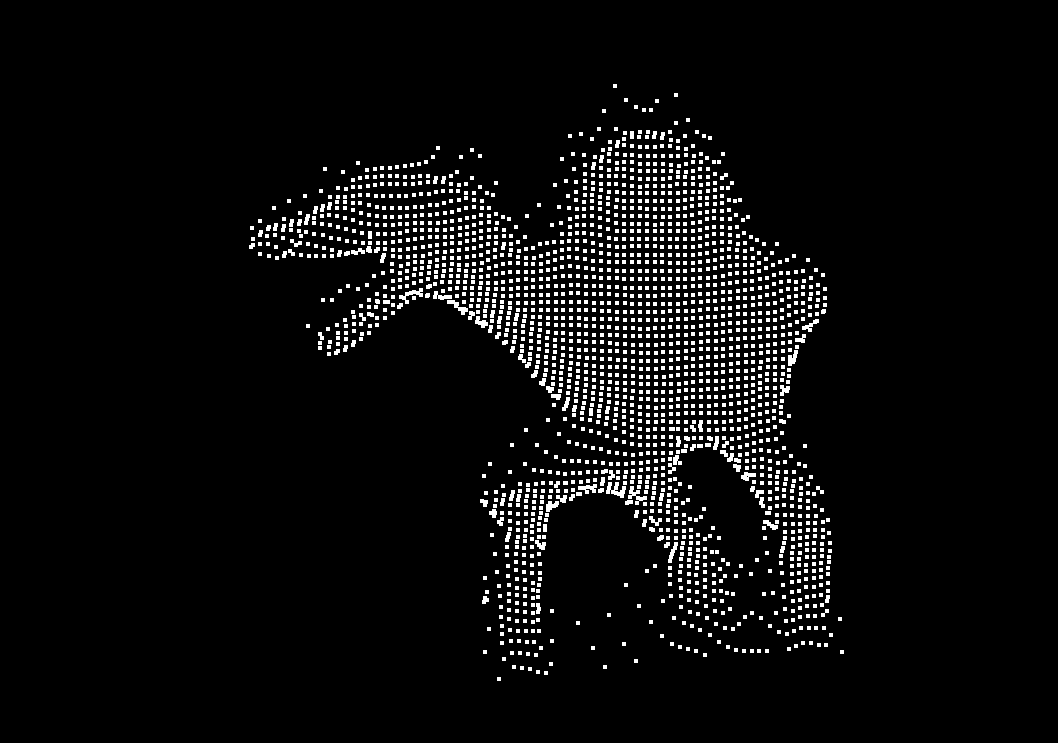}
    \\
    \caption{Depth maps estimated by ZoeDepth and 3D points.} 
    \label{fig:ZoeDepth_depth_and_3dpoint}
\end{figure}

Please refer to Fig.~\ref{fig:ZoeDepth_depth_and_3dpoint} for the visualization of depth maps estimated by ZoeDepth and 3D points on our test set.

\begin{IEEEbiography}[{\includegraphics[width=1in,height=1.25in,clip,keepaspectratio]{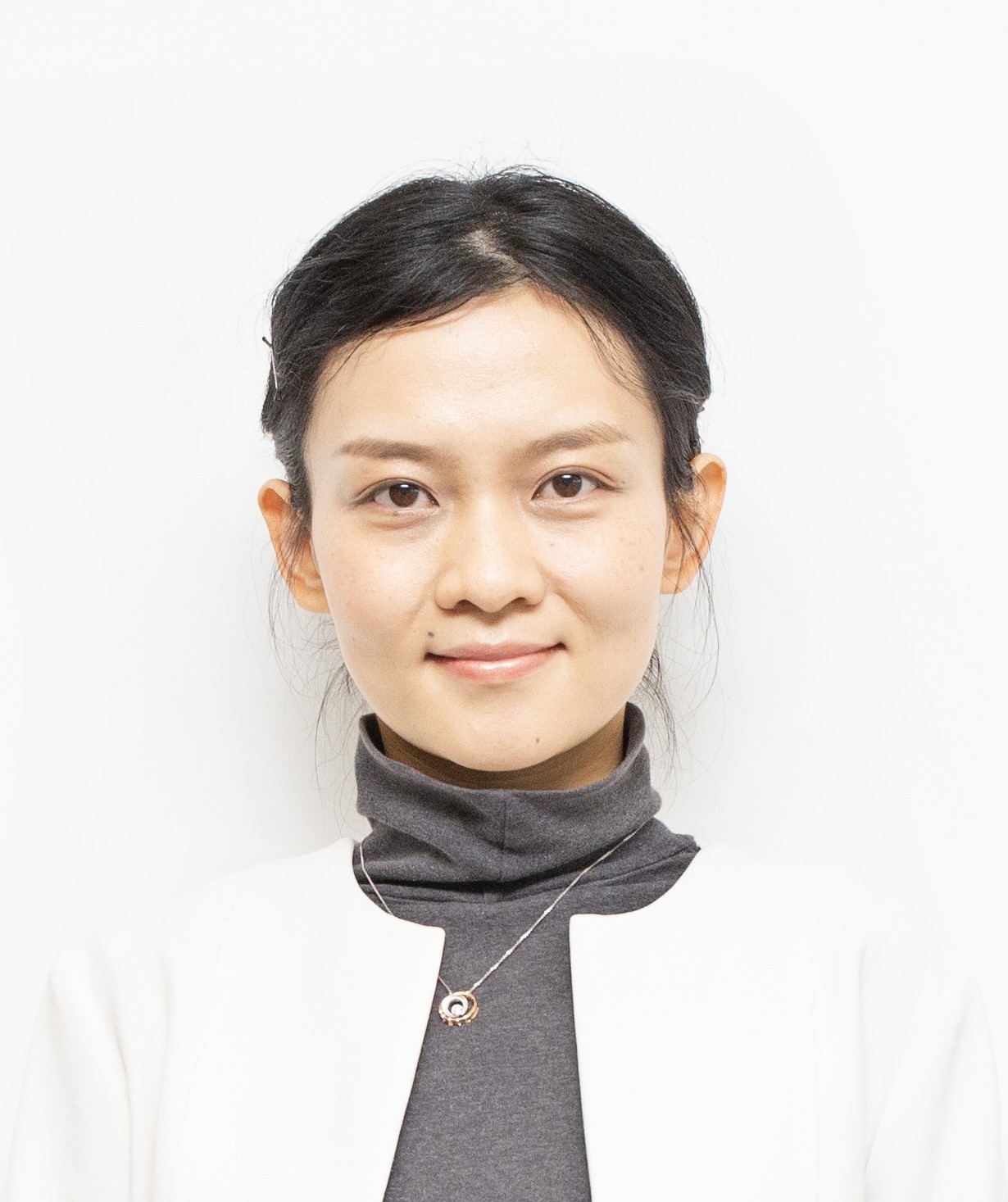}}]{Yujing Sun}
is currently a research assistant professor at the University of Hong Kong. Before that, she received a bachelor's degree from University of Minnesota, Twin Cities in 2013 and a PhD. in Computer
Science from the University of Hong Kong in 2018, under the supervision of Prof.Wenping Wang. Her research interests include financial data analysis, computer graphics, image processing, and biometrics.
\end{IEEEbiography}

\begin{IEEEbiography}[{\includegraphics[width=1in,height=1.25in,clip,keepaspectratio]{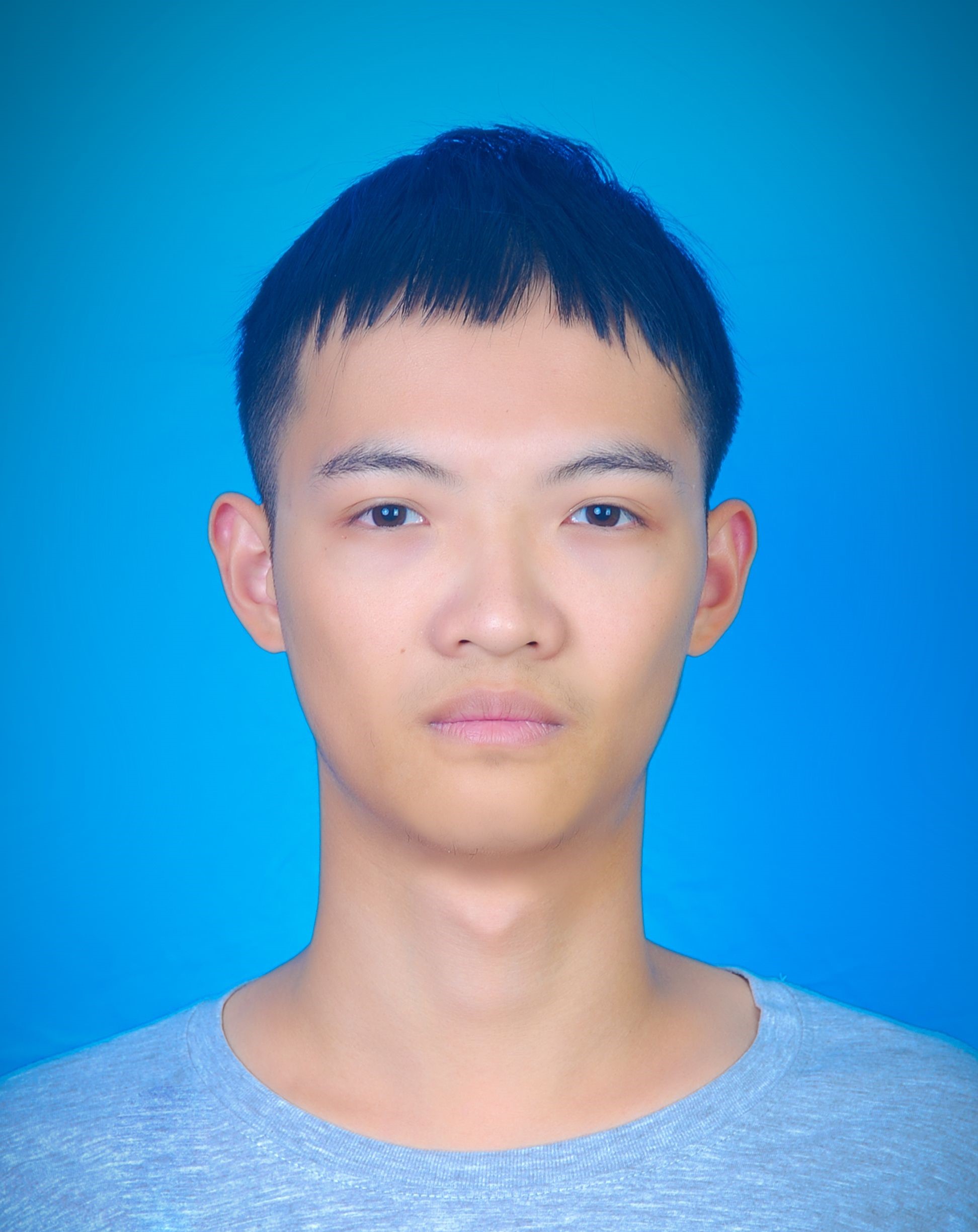}}]{Caiyi Sun}
Caiyi Sun is currently pursuing the bachelor’s degree with the School of Integrated Circuits and Electronics, Beijing Institute of Technology, Beijing, China. His research interests include 3D computer vision and financial data analysis.
\end{IEEEbiography}

\begin{IEEEbiography}[{\includegraphics[width=1in,height=1.25in,clip,keepaspectratio]{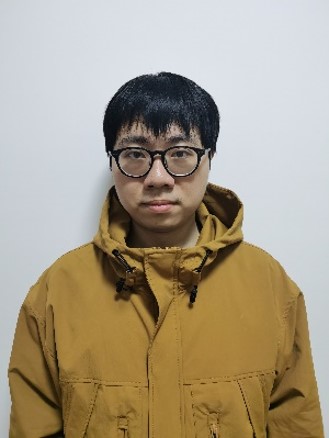}}]{Yuan Liu}
Yuan Liu is a final-year PhD student at HKU advised by Prof. Wenping Wang and Prof. Taku Komura. In 2018-2019, he worked in the CAD\&CG lab of Zhejiang University advised by Prof. Xiaowei Zhou. He received a master's degree at LIESMARS of Wuhan University advised by Prof. Bisheng Yang. His research interest includes computer graphics and 3D computer vision.
\end{IEEEbiography}

\begin{IEEEbiography}
[{\includegraphics[width=1in,height=1.25in,clip,keepaspectratio]{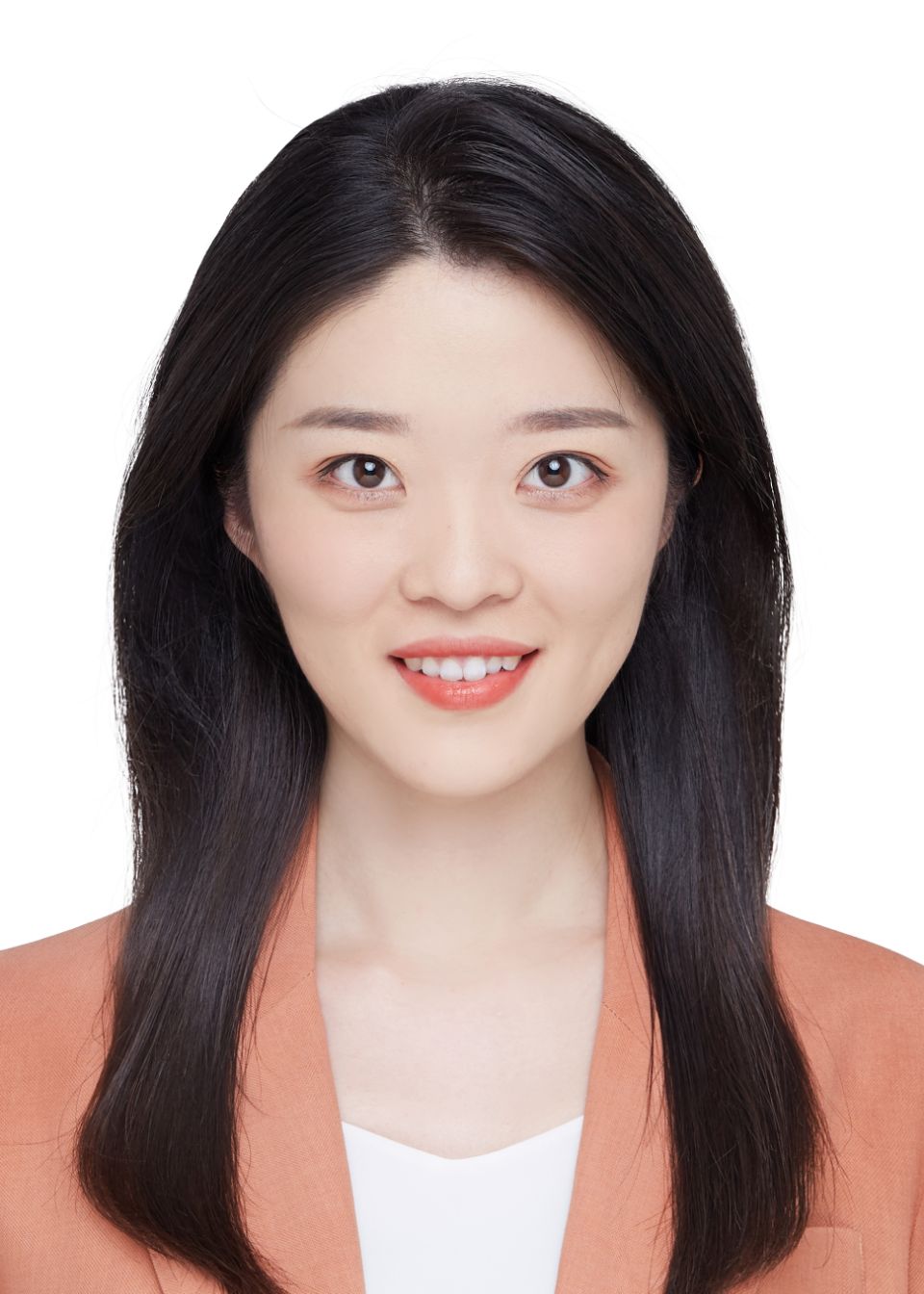}}]{Yuexin Ma}
received the PhD degree from the Department of Computer Science at University of Hong Kong in 2019.
Now she is an assistant professor in ShanghaiTech university, China, leading 4DV Lab. Her research interests lie on the computer vision and deep learning. Particularly she is interested in 3D scene understanding, multi-modal perception, human-machine cooperation, and autonomous driving. She has published dozens of papers on TPAMI, IJCV, CVPR, ICCV, ECCV, IJCAI, AAAI, etc.
\end{IEEEbiography}

\begin{IEEEbiography}[{\includegraphics[width=1in,height=1.25in,clip,keepaspectratio]{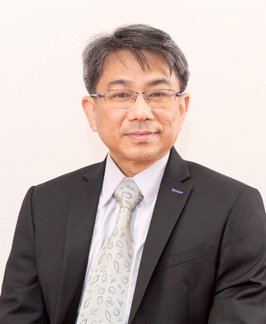}}]{Siu Ming Yiu} 
is a full professor and the deputy head of the department of computer science, the University of Hong Kong. He received his PhD in Computer Science from the University of Hong Kong. Prof. Yiu has Published 100+ papers in referred journals and conferences (Citations (8627), h-index (41), i10-index (98)) and is Conference/programme chairs in prestigious conferences in both areas of cryptography and bioinformatics. 
\end{IEEEbiography}
\vfill

\end{document}